\documentclass{article}

\usepackage{styles/arxiv}
\usepackage{styles/arxiv_abbreviations}
\usepackage[utf8]{inputenc} 
\usepackage[T1]{fontenc}    
\usepackage{hyperref}       
\usepackage{url}            
\usepackage{booktabs}       
\usepackage{amsfonts}       
\usepackage{nicefrac}       
\usepackage{microtype}      
\usepackage{xcolor}         
\usepackage{array}
\usepackage{float}
\usepackage{graphicx} 
\usepackage{amsmath}
\usepackage{amsfonts}
\usepackage[nonumberlist, nogroupskip, nolist, nopostdot, acronym]{glossaries}
\usepackage{adjustbox}
\usepackage{xspace}
\usepackage{booktabs}
\usepackage{multicol}
\usepackage{multirow} 
\usepackage[capitalize]{cleveref} 
\usepackage{pifont}

\newcommand{\cmark}{\ding{51}}
\newcommand{\xmark}{\ding{55}}

\usepackage{tcolorbox}
\usepackage{adjustbox}
\usepackage{listings}
\crefname{section}{Sec.}{Secs.}
\Crefname{section}{Section}{Sections}
\crefname{table}{Tab.}{Tabs.}
\Crefname{table}{Table}{Tables}

\newacronym{ai}{AI}{Artificial Intelligence}
\newacronym{amax}{ActMax}{Activation Maximization}
\newacronym{ch}{CH}{Clever Hans}
\newacronym{cnn}{CNN}{Convolutional Neural Network}
\newacronym{nn}{NN}{Neural Network}
\newacronym{crc}{CRP}{Concept Relevance Propagation}
\newacronym{dl}{DL}{Deep Learning}
\newacronym{dnn}{DNN}{Deep Neural Network}
\newacronym{ga}{GA}{Gradient Ascent}
\newacronym{gan}{GAN}{Generative Adversarial Network}
\newacronym{hitl}{HITL}{Human in the Loop}
\newacronym{lrp}{LRP}{Layer-wise Relevance Propagation}
\newacronym{ml}{ML}{Machine Learning}
\newacronym{mlp}{MLP}{Multilayer Perceptron}
\newacronym{llm}{LLM}{Large Language Model}
\newacronym{rmax}{RelMax}{Relevance Maximization}
\newacronym{rnn}{RNN}{Recurrent Neural Network}
\newacronym{tcav}{TCAV}{Testing With Activation Vectors}
\newacronym{xai}{XAI}{eXplainable Artificial Intelligence}
\newacronym{vit}{ViT}{Vision Transformer}

\newacronym{lll}{LLL}{Low-Level hidden Layers}
\newacronym{mll}{MLL}{Mid-Level hidden Layers}
\newacronym{hll}{HLL}{High-Level hidden Layers}
\newacronym{fcl}{FCL}{Fully-Connected Layers}
\newacronym{mag}{MAG}{Magnitude Flag}
\newacronym{sae}{SAE}{Sparse Auto Encoder}
\newacronym{cbm}{CBM}{Concept Bottle-neck Model}

\newacronym{sem}{SEM}{Standard Error of Mean}
\newacronym{eap}{EAP}{Edge Attribution Patching}
\newacronym{acdc}{ACDC}{Automated Circuit DisCovery}
\newacronym{dag}{DAG}{Directed Acyclic Graph}
\newacronym{ioi}{IOI}{Indirect Object Identification}
\newacronym{cav}{CAV}{Concept Activation Vector}
\newacronym{ifr}{IFR}{Information Flow Routes}
\newacronym{rur}{RUR}{Response Uniqueness Ratio}

\usepackage{xspace}
\glsdisablehyper

\hypersetup{
    colorlinks,
    linkcolor={red!60!black},
    citecolor={blue!70!black},
    urlcolor={blue!70!black}
}

\definecolor{myred}{HTML}{C23F40}
\definecolor{myAcronymColor}{RGB}{0, 128, 128} 

\let\oldgls\gls
\renewcommand{\gls}[1]{\textcolor{black}{\oldgls{#1}}}

\usepackage{afterpage}

\title{Attribution-Guided Pruning for Insight and Control: Circuit Discovery and Targeted Correction in Small-scale LLMs}

\author{%
  \And {Sayed Mohammad Vakilzadeh Hatefi$^{1,2,\dagger}$~~~~}{} \And
  {Maximilian Dreyer$^{2}$~~~~}{} \And
  {Reduan Achtibat$^{2}$~~~~}{} \And 
  {Patrick Kahardipraja$^{2}$~~~~}{} \And
  {Thomas Wiegand$^{2,3,4}$~~~~}{} \And
  {Wojciech Samek$^{2,3,4,\dagger}$~~~~}{} \AND
  {Alexander Binder$^{1}$~~~~}{}
  {Sebastian Lapuschkin$^{2,5,\dagger}$~~~~}{}
  \\
  \\
  $^{1}$Data Science Center ScaDS.AI, Universit{\"a}t Leipzig\\
  $^{2}$Department of Artificial Intelligence, Fraunhofer Heinrich-Hertz-Institute\\
  $^{3}$Department of Electrical Engineering and Computer Science, Technische Universit{\"a}t Berlin \\
  $^{4}$BIFOLD - Berlin Institute for the Foundations of Learning and Data\\
  $^{5}$Centre of eXplainable Artificial Intelligence, Technological University Dublin\\
  $^\dagger$ corresponding authors: \texttt{mohammad.hatefi@uni-leipzig.de}\\ \texttt{\{wojciech.samek,sebastian.lapuschkin\}@hhi.fraunhofer.de}
}

\begin{document}

\maketitle

\begin{abstract}
\glspl{llm} are widely deployed in real-world applications, yet their internal mechanisms remain difficult to interpret and control, limiting our ability to diagnose and correct undesirable behaviors. Mechanistic interpretability addresses this challenge by identifying circuits -- subsets of model components responsible for specific behaviors. However, discovering such circuits in LLMs remains difficult due to their scale and complexity. We frame circuit discovery as identifying parameters that contribute most to model outputs on task-specific inputs, and use \gls{lrp} with reference samples to attribute and extract these components via pruning. Building on this, we introduce contrastive relevance to isolate circuits associated with undesired behaviors while preserving general capabilities, enabling targeted model correction. On OPT-125M, we show that pruning as little as $\sim$0.3\% of neurons substantially reduces toxic outputs, while pruning approximately 0.03\% of weight elements mitigates repetitive text generation without degrading general performance. These results establish attribution-guided pruning as an effective mechanism for identifying and intervening on behavior-specific circuits in LLMs. We further validate our findings on additional small-scale language models, demonstrating that the proposed approach transfers across architectures.
Our code is publicly available at \url{https://github.com/erfanhatefi/SparC3}.
\end{abstract}

\section{Introduction}

Since the introduction of the Transformer architecture \cite{vaswani2017attention}, language modeling has undergone a paradigm shift, enabling the development of increasingly capable models across a wide range of scale. Despite their success, the internal mechanisms underlying their behaviors remain difficult to understand, limiting transparency and raising concerns about reliability and trustworthiness in sensitive applications \cite{bender2021dangers, liu2023trustworthy}. This has motivated growing research in eXplainable AI (XAI) and mechanistic interpretability, which aims to analyze internal computations of neural networks. Within this line of work, circuit discovery \cite{conmy2023towards, ferrando2024information, marks2024sparse} seeks to identify subsets of model components responsible for specific behaviors, ranging from simple linguistic operations to undesired behaviors such as toxic generation. While modern \glspl{llm} are increasingly capable, their scale makes mechanistic analysis computationally challenging. We therefore focus on \emph{small-scale models} (\eg OPT-125M, TinyLlama, Qwen2-0.5B), which provide a controlled and tractable setting to study behavior-specific circuits. Small-scale models are also practically relevant for resource-constrained and domain-specific deployments. \textbf{We hypothesize that insights from small-scale models transfer to larger architectures}, supported by prior work showing that circuit-level algorithms and functional components remain consistent across model scales, even when their specific implementations differ \cite{tigges2024llm}. While our experiments are conducted on small-scale models, we believe our work is not restricted to model size and is directly applicable to larger architectures.

\begin{figure}[t]
  \centering
    \includegraphics[width=0.99\linewidth]{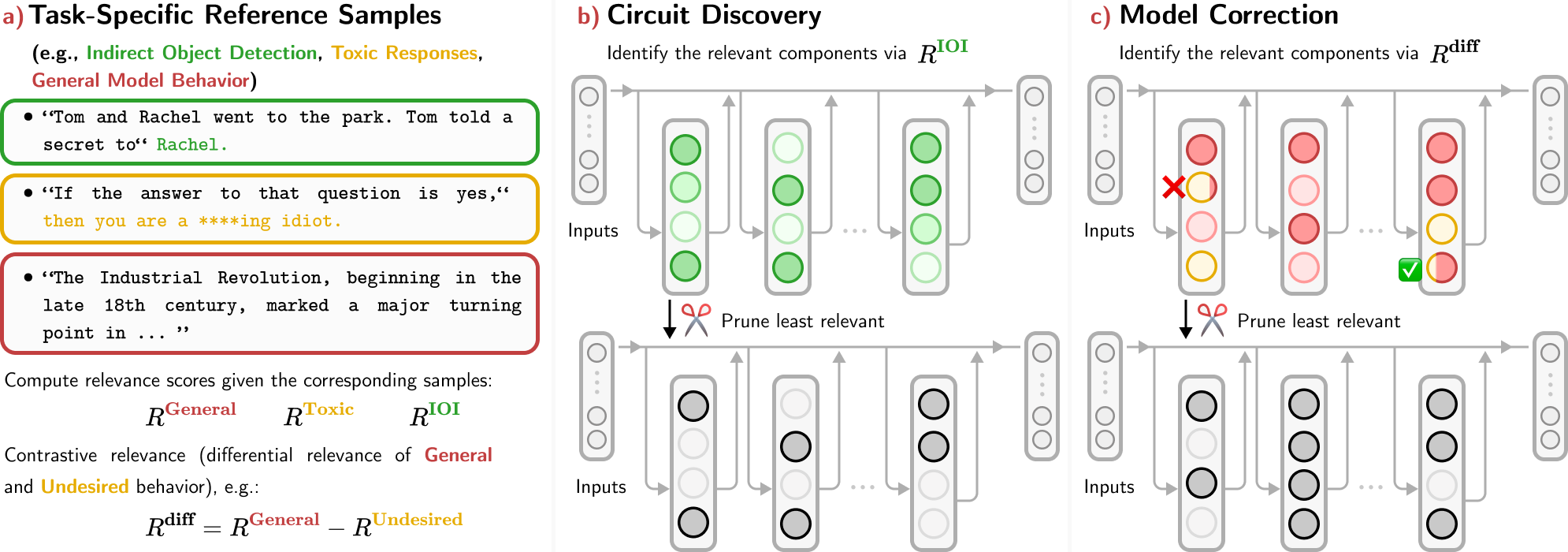}
    \caption{
    Overview of attribution-guided circuit discovery and model correction. 
    \textbf{\textcolor{myred}{a)}} Attribution methods (e.g., LRP) compute relevance scores $R$ for model components using task-specific reference samples. 
    \textbf{\textcolor{myred}{b)}}
    Relevance scores localize the subgraph responsible for a task. Pruning low-relevance components isolates a sparse circuit that preserves the target behavior.
    \textbf{\textcolor{myred}{c)}} Contrastive relevance ($R^{\text{diff}} = R^{\text{General}} - R^{\text{Undesired}}$) identifies harmful components, whose removal mitigates undesired behavior while preserving general capabilities.
    }
   \label{fig:introduction} 
\end{figure}

Existing circuit discovery approaches \cite{olah2020zoom, wang2019structured, conmy2023towards, ferrando2024information} typically identify causal subgraphs via activation interventions, tracing information flow and token-level interactions across layers. While these methods often aggregate intervention effects over sets of related inputs to extract task-level circuits, they operate on activation-level representations tied to specific forward passes. As a result, the circuits are expressed in terms of activation patterns and computational paths, making direct manipulation at the parameter level non-trivial. In contrast, we define circuits as behavior-specific subsets of model parameters and architectural components (\eg attention heads, neurons, and weights) that collectively contribute to a behavior across related inputs. This parameter-space view enables efficient discovery and direct manipulation of identified components (\eg via pruning). Under this formulation, circuits can be identified using attribution methods. We employ \gls{lrp} \cite{bach2015pixel, montavon2019layer, pmlr-v235-achtibat24a}, a state-of-the-art attribution method widely used for neural networks, including transformers, which assigns relevance scores to model components based on their contribution to the model’s output. Using behavior-specific prompts, we compute relevance scores and apply attribution-guided pruning \cite{hatefi2024pruning, yeom2021pruning} to extract sparse circuits (see \cref{fig:introduction}).

Beyond circuit discovery, our framework naturally enables targeted model correction. Given prompts that elicit an undesired behavior, attribution scores identify the corresponding behavior-specific components. However, some components may also contribute to general capabilities (\eg linguistic structure or grammaticality). To address this, we contrast attribution scores obtained from undesired prompts with those from general prompts -- representing normal model usage -- thereby isolating components specific to the undesired behavior. This enables targeted mitigation while preserving overall model performance.

We propose a unified framework for attribution-guided pruning of language models, supporting two primary applications: 1) \textbf{Circuit discovery}, where we extract fine-grained parameter-level subgraphs responsible for specific behaviors (\eg indirect object identification); and 2) \textbf{Model correction}, where we identify and remove circuits associated with undesired behaviors while preserving overall model performance. Table~\ref{tab:framework_comparison} compares our approach to prior work, highlighting that existing methods typically address only one of these objectives, and that attribution quality determines the effectiveness of pruning-based circuit discovery and targeted model correction. While not our primary focus, model compression arises as a direct byproduct of the same framework.

\begin{table}[b]
\centering
\small
\setlength{\tabcolsep}{3pt}
\caption{
    Comparison of methods across circuit discovery, model correction, compression, and parameter-level attribution. Prior approaches typically address only a subset of these capabilities, whereas our attribution-guided pruning framework unifies them. $\sim$ denotes partial capability.
}
\label{tab:framework_comparison}
\begin{tabular}{lcccc}
\toprule
Method & Circuit Discovery  & Model Correction  & Parameter-level & Compression \\
\midrule
ACDC, EAP (Activation-based) & \cmark & \xmark & \xmark & \xmark \\
Gradient (Saliency) & $\sim$ & $\sim$ & \cmark & $\sim$ \\
Wanda (Pruning-based) & $\sim$ & $\sim$ & \cmark & \cmark  \\
\midrule
LRP-guided Pruning (ours) & \textbf{\cmark} & \textbf{\cmark} & \textbf{\cmark} & \textbf{\cmark} \\
\bottomrule
\end{tabular}
\end{table}


\section{Related work}

\paragraph{Attribution methods}
\label{subsec:attribution}
Attribution methods explain neural network predictions by assigning importance scores to inputs or internal components. Common approaches include gradient-based saliency methods \cite{smilkov2017smoothgrad}, Integrated Gradients \cite{sundararajan2017axiomatic}, and DeepLIFT \cite{shrikumar2017learning} which estimate how changes in inputs affect model outputs. \gls{lrp} \cite{bach2015pixel, montavon2019layer} instead redistributes the prediction scores backward through the network via relevance conservation, producing importance scores for intermediate neurons and parameters. Compared to gradient-based methods which suffer from gradient shattering \cite{balduzzi2017shattered}, \gls{lrp} yields more stable and faithful attributions \cite{pmlr-v235-achtibat24a, hatefi2024pruning, yeom2021pruning}, and has recently been extended for transformers to reveal meaningful computation paths \cite{pmlr-v235-achtibat24a, ali2022xai}. In this work, we leverage \gls{lrp} and its transformer-specific variant (AttnLRP \cite{pmlr-v235-achtibat24a}) to identify task-relevant subnetworks in \glspl{llm}.


\paragraph{Pruning and compression}
\label{subsec:pruning}
Pruning reduces model size by removing parameters with limited contribution to performance. Recent methods for \glspl{llm} rely on heuristics such as weight magnitude or activation-based importance. For example, Wanda \cite{sun2023simple} prunes weights based on their magnitudes scaled by activation norms. Attribution-based pruning has also been explored, where attribution scores are used to identify important components \cite{yeom2021pruning, hatefi2024pruning}. In this work, we extend LRP-based pruning to \glspl{llm} using attribution to identify task-relevant parameters. While our focus is circuit discovery and model correction, we additionally show competitive compression performance (\cref{app:unstructured_pruning}).


\paragraph{Circuit discovery}
\label{subsec:mechinterp}
Circuit discovery aims to identify internal components, such as attention heads and neurons from \gls{mlp}, that drive specific model behaviors. Most existing approaches operate on activation-level representations, identifying circuits as causal subgraphs of token-level interactions for specific inputs. Methods such as \glspl{sae} \cite{marks2024sparse} learn interpretable activation features but require additional training, while activation patching approaches such as \gls{acdc} \cite{conmy2023towards} estimate causal importance via systematic interventions. Although effective, these intervention-based approaches are computationally expensive and typically extract circuits using attribution-score thresholds. More efficient variants, including \gls{ifr} \cite{ferrando2024information} and \gls{eap} \cite{syed2023attribution}, approximate these interventions using gradient-based or attribution-based signals. More recently, Relevance Patching (RelP) \cite{jafari2025relp} integrates \gls{lrp} into the patching framework, but remains within the activation-level paradigm. In contrast, we adopt a parameter-level perspective: we identify circuits as subsets of model parameters contributing to a behavior across a set of inputs. We use \gls{lrp} to rank parameters by relevance and extract circuits via pruning, enabling efficient identification of sparse, behavior-specific subnetworks and directly connecting circuit discovery with model correction and compression.


\paragraph{Model correction}
\label{subsec:model_correction}
\glspl{dnn} often exhibit undesirable behaviors, such as biased predictions or toxic outputs. While data cleaning or fine-tuning can mitigate these issues, such solutions are typically expensive and impractical at scale. Existing methods address this in different ways. In vision models, \cite{ross2017right} incorporate attribution into training objectives, while \cite{schramowski2020making, anders2022finding, pahde2023reveal, dreyer2024hope} mitigate biases by manipulating latent representations. For \glspl{llm}, methods such as \cite{ravfogel2020null, turner2023steering} steer behavior via directions in latent space, but do not modify model parameters directly or provide compression benefits. Other approaches rely on fine-tuning for alignment \cite{ouyang2022training}, or identify knowledge neurons using gradients for targeted editing \cite{dai2021knowledge}. Recent model editing methods, such as ROME \cite{meng2022locating} and MEMIT \cite{meng2022mass}, modify specific factual knowledge via targeted weight updates. In contrast, circuit discovery provides a natural mechanism for targeted model correction by identifying the components responsible for specific behaviors. In this work, we leverage \gls{lrp} to localize the components associated with the undesirable behaviors. By contrasting relevance from harmful and benign reference samples, we isolate and prune these parameters, enabling targeted correction without fine-tuning while preserving overall performance.


\section{Methods}

We introduce a framework for attribution-guided pruning that connects component-level scoring with circuit discovery and model correction. This framework can be instantiated with \gls{lrp} (\cref{sec:methods:lrp}) or importance-based pruning approaches (\eg Wanda, \cref{app:wanda}).


\subsection{Attribution-based pruning}
\label{sec:methods:attribution_pruning}

Building on the framework introduced by \cite{yeom2021pruning, hatefi2024pruning}, let $ \Psi = \{\psi_{1}, \dots, \psi_{p}\}$ denote a set of $p$ components -- defined as structural units like neurons from \glspl{mlp}, attention heads, or other trainable parameters -- that constitute a \gls{dnn}, and let $\mathcal{X}_\text{ref} = \{x_1, x_2, \dots, x_{n_{\text{{ref}}}}\}$ represent a set of reference samples. For each component $\psi_{k} \in \Psi$ and reference sample $x_i \in \mathcal{X}_\text{ref}$, we define $R_{\psi_{k}}({x_i})$ as the relevance
(or importance) score obtained from an attribution method (\ie \gls{lrp}). By aggregating these scores across all reference samples and applying the normalization described in \cref{eq:compute_importance}, we obtain $\mathcal{R} = \{\bar{R}_{\psi_{1}}, \bar{R}_{\psi_{2}}, \dots, \bar{R}_{\psi_{p}}\}$, the set of normalized relevance scores for all components.
 
\begin{equation}
    \label{eq:compute_importance}
    \bar{R}_{\psi_{k}} = \frac{1}{n_{\text{ref}}}\sum_{i=1}^{n_{\text{ref}}} R_{\psi_{k}}({x_i})\,.
\end{equation}

Regardless of the pruning approach, whether it is structured, fully unstructured, per-layer unstructured, or row-wise unstructured (an overview of these approaches is given in \cref{app:prunig_approaches}), we can order components based on their attributed relevance scores. Specifically, we assign indices $c$ by sorting components in ascending order of relevance, up~to the $q$-th~place:

\begin{equation}
\label{method:attribution_pruning:sorting}
    \centering
    \{c\}_{q} = \text{argsort}(\mathcal{R})_{1, 2,\dots, q}\,.
\end{equation}

Defining $\textbf{1}$ to represent an indicator function with condition $i \in \{c\}_{q}$, the $q$ least relevant components can be pruned by masking as:

\begin{equation}
    \centering
    \forall {\psi_{i} \in \Psi}: \psi_{i} \mapsto (1- \textbf{1}_{i \in \{c\}_{q}}) \psi_{i}\,.
\end{equation}


\subsection{Layer-wise Relevance Propagation}
\label{sec:methods:lrp}

\glsdesc{lrp}~\cite{bach2015pixel, montavon2019layer} treats a neural network with $L$ layers as a \gls{dag}, such that for a given input $x$:

\begin{equation}
    \centering
    f(x) = f^{L} \circ \dots \circ f^{l} \circ f^{l-1} \circ \dots \circ f^{1} (x)\,.
\end{equation}

\gls{lrp} employs a backpropagation process via specific rules designed to allocate ``relevance'' scores to (both parametric and non-parametric) edges of the \gls{dag}, proportional to their contribution to the final prediction. At first, this process begins at the last layer $f^{L}$ by initializing the relevance score of $R_j^L$ at output neuron $j$ of $f^{L}$, using the maximum logit at the final token position, and ultimately redistributing this score to its input variables. To elaborate the redistribution at a specific layer $l$, denote  $z_{ij}$  to be the mappings of input neuron $i$ to output neuron $j$ with output $z_j$ (pre-activation) which in linear layers this notation is represented by $z_{ij} = a_i w_{ij}$ with  $w_{ij}$ as the weight parameters and $a_i$ as the activation of neuron $i$. \gls{lrp} then redistributes the upper layer relevance quantity of $R_j^l$ towards the lower layers proportionally to the relative contributions of $z_{ij}$ to $z_j$, resulting in $ R_{i\leftarrow j}^{(l-1, l)}$ that quantifies the contribution of neuron $i$ at layer $l-1$, to the activation of neuron $j$ at layer $l$: 

\begin{equation} \label{eq:lrp_basic}
    \centering
    R_{i\leftarrow j}^{(l-1, l)} = \frac{z_{ij}}{z_{j}} R_{j}^l\,.
\end{equation}

An aggregation of all $R_{i\leftarrow j}^{(l-1, l)}$ obtains the contribution of neuron $i$ to all upper layer neurons $j$:
\begin{equation}
\label{lrp:conservation}
      \sum_i R^{l-1}_i = \sum_{i,j} R^{(l-1, l)}_{i\leftarrow j} = \sum_j R^l_j 
\end{equation}

Extra steps on obtaining relevance scores from attention heads, and scores of each weight parameter, are discussed in detail at \cref{app:lrp_in_detail}.


\subsection{Circuit discovery}
\label{sec:methods:circuit_discovery}

We define a circuit as a subnetwork comprising a subset of model components $\mathcal{C} \subseteq \Psi$, where $\Psi$ denotes the set of candidate components under analysis (e.g., weights, neurons, or attention heads). The circuit $\mathcal{C}$ corresponds to components whose relevance scores account for most attribution to a target behavior. In practice, we extract this circuit by progressively pruning components $\psi_i \in \Psi$ with low attribution scores computed on reference samples $\mathcal{X}_\text{ref}$ designed to capture the behavior of interest. We evaluate the model across a range of sparsity levels obtained through this pruning process (see \cref{fig:circuit_discovery}) using a task-specific performance metric (\eg IOI accuracy). The final circuit $\mathcal{C}$ is defined as the sparsest subnetwork that preserves the target behavior within a predefined tolerance relative to the original model.

In contrast, existing methods (\eg \cite{conmy2023towards, syed2023attribution, ferrando2024information}) typically define circuits as computational subgraphs derived from hidden activations across tokens, capturing information flow for specific inputs. While these approaches can aggregate intervention effects across related inputs to extract task-level circuits, the resulting circuits remain expressed as activation flows and computational paths, making direct parameter-level manipulation difficult (\eg via pruning). Our approach instead identifies circuits directly from model components via their relevance scores. Pruning thus serves as the extraction mechanism, yielding input-independent circuits that are easier to interpret and more practical for correcting unwanted behaviors, analysis, and compression.


\subsection{Model correction}
\label{sec:methods:model_correction}

Let $\mathcal{X}_{\text{ref}}^{\text{General}}$ and $\mathcal{X}_{\text{ref}}^{\text{Undesired}}$ denote the sets of reference samples that capture the model’s general behavior (\eg datasets such as Wikipedia and C4) and a specific undesired behavior (\eg toxicity), respectively. Applying the framework described in \cref{sec:methods:attribution_pruning} to each of these sets yields two attribution scores $\bar{R}^{\text{General}}$ and $\bar{R}^{\text{Undesired}}$. To isolate components responsible for the undesired behavior, we contrast attribution scores obtained from undesired samples $\bar{R}^{\text{Undesired}}$ with those obtained from general-purpose data $\bar{R}^{\text{General}}$:

\begin{equation}
\label{model_correction:differential_attribution}
\bar{R}^{\text{diff}}_{\psi_{k}} =  \bar{R}^{\text{General}}_{\psi_{k}} -  \bar{R}^{\text{Undesired}}_{\psi_{k}}
\end{equation}

Intuitively, this contrasts components that support the model’s general functionality with those that contribute to the undesired behavior. The resulting scores form the differential attribution set $\mathcal{R}^\text{diff} = \{\bar{R}^\text{diff}_{\psi_{1}}, \bar{R}^\text{diff}_{\psi_{2}}, \dots, \bar{R}^\text{diff}_{\psi_{p}}\}$, which is subsequently used for pruning. Following the pruning procedure from \cref{method:attribution_pruning:sorting}, we sort $\mathcal{R}^\text{diff}$ in ascending order so that components with low (or negative) differential relevance -- those contributing more strongly to the undesired behavior than to the model’s general functionality -- are removed first.


\section{Experiments}

Our experiments evaluate the proposed framework on circuit discovery and model correction (see \cref{fig:introduction} for an overview). We focus our main analysis on OPT-125M, a widely used model in mechanistic interpretability. We evaluate TinyLlama and Qwen2-0.5B in \cref{app:sec:other_models} to assess generalizability across small-scale architectures. We primarily compare against Wanda \cite{sun2023simple}, a strong pruning baseline for general function preservation that attributes parameters using weight magnitudes and activations on reference samples, and has shown strong performance in LLM compression~\footnote{Additional compression results demonstrating the competitiveness of our framework are provided in \cref{app:unstructured_pruning}.}.
Its reliance on reference samples makes it suitable for comparison in our setting. In addition, we include gradient-based attribution as a widely used baseline, and neuron activations (for structured pruning) to further assess attribution quality and support our choice of \gls{lrp}. We evaluate multiple pruning strategies, including both structured and unstructured approaches (see \cref{app:prunig_approaches} for details).


\subsection{Discovering task-specific and sparse circuits}
\label{exp:circuit_discovery}
Understanding how specific behaviors are implemented within a model requires identifying sparse subgraphs -- so-called circuits -- that are necessary and sufficient for a given task. We evaluate our framework on Indirect Object Identification (IOI) task~\cite{conmy2023towards}, where the model must identify the correct indirect object in a sentence. This task serves as a standard benchmark for circuit discovery due to its well-defined structure and known localization. We test whether attribution-guided pruning via \gls{lrp} can recover circuits that preserve task performance while achieving high sparsity, \ie by removing irrelevant components without degrading behavior.


\paragraph{Experimental settings}
We use the 125M-parameter OPT model~\cite{zhang2022opt} and generate six reference sets of 128 IOI-like sequences, following~\cite{conmy2023towards} with different random seeds. To extract circuits, we compare \gls{lrp} and Wanda-based pruning and include gradient-based attribution and neuron activations as baselines, following prior circuit discovery works~\cite{ferrando2024information, conmy2023towards}.

Intervention-based methods (\eg ACDC, IFR, RelP, EAP) operate on activation-level causal graphs capturing token-level information flow, whereas our approach targets component-level scoring in parameter space, enabling direct manipulation via pruning. We evaluate all methods at two levels of granularity: \textbf{1}) \emph{structured pruning}, where entire neurons or attention heads are removed, and \textbf{2}) \emph{unstructured pruning}, where individual weight elements -- edges between neurons ~--~are~pruned based on their attributed relevance.

A circuit is considered high-quality if it \textbf{i}) includes all task-critical components -- whose removal significantly degrades performance -- and \textbf{ii}) excludes irrelevant ones. We assess this via performance-sparsity curves, measuring task accuracy across a range of pruning rates. Inspired by feature perturbation tests for attribution evaluation \cite{samek2016evaluating}, these curves characterize how resilient a circuit is to pruning: a flat or increasing trend indicates redundancy, while sharp performance drops signal the removal of essential components.

We further analyze how attribution scores are distributed across components and layers for the IOI task, focusing on \gls{lrp} and Wanda as representative attribution-based pruning methods. We concentrate on the \texttt{fc1} layers, which exhibit the clearest localization patterns and are central to our correction experiments. This analysis reveals whether a method identifies compact multi-layer circuits or instead concentrates importance more diffusely or predominantly in later layers. Results for \texttt{fc2} are provided in \cref{app:localzation_fc2}.

\begin{figure}
  \centering
  \includegraphics[width=0.99\linewidth]{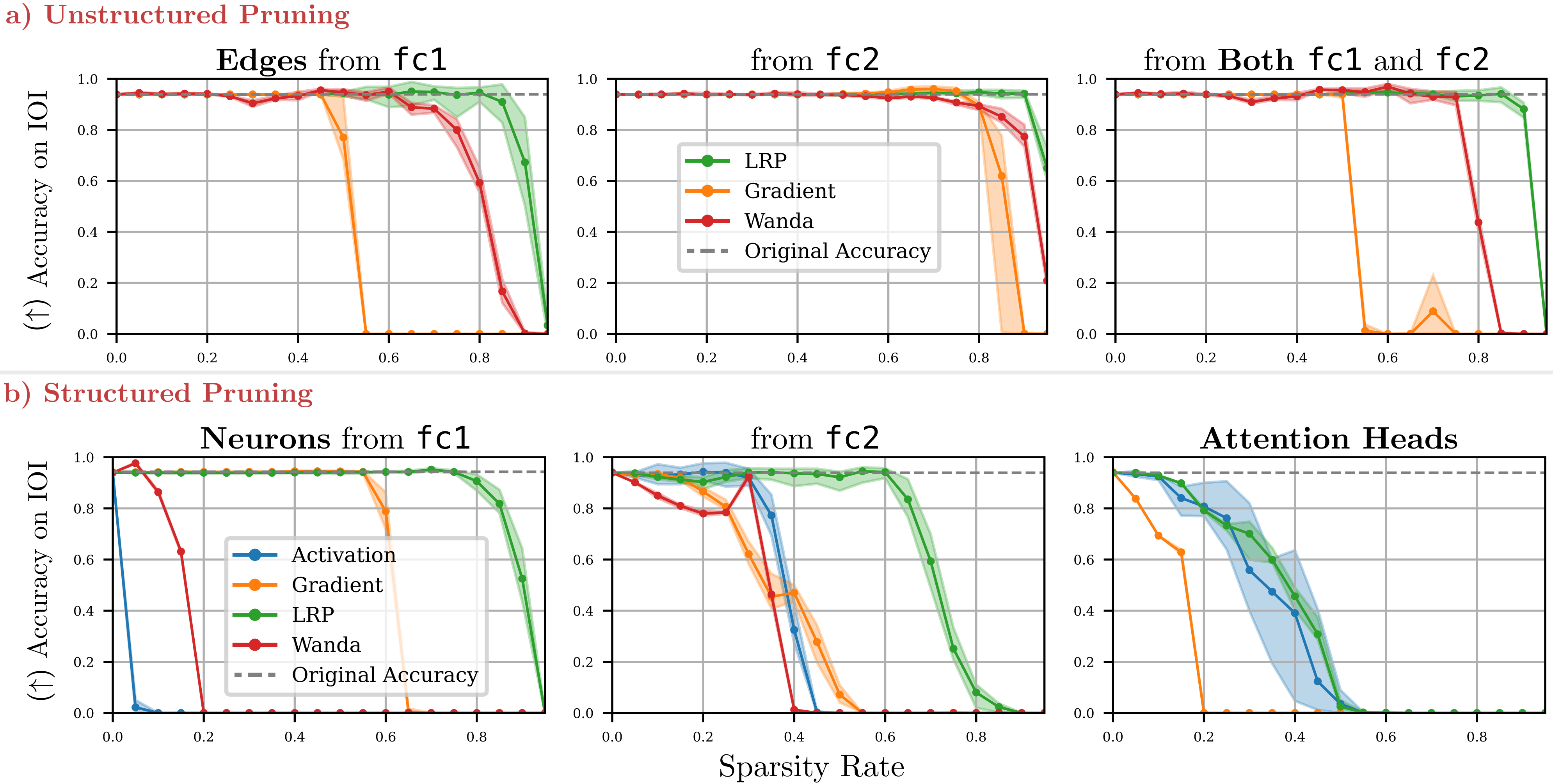}
  \caption{
  \emph{\textcolor{myred}{a)}} Parameter-level IOI circuits extracted via unstructured pruning in \texttt{fc1} and \texttt{fc2} layers of OPT-125M. \gls{lrp} and Wanda yield substantially sparser circuits than gradient-based attribution and neuron activations (all methods shown at optimal settings; see \cref{app:circuit_discovery_unstrcutred} for complete detail). Wanda uses row-wise pruning, while \gls{lrp} and gradient apply global sorting across layers. \emph{\textcolor{myred}{b)}} Structured pruning of neurons and attention heads yields less sparse circuits across methods. The shaded region shows the mean $\pm$ standard deviation.
  }
  \label{fig:circuit_discovery}
\end{figure}

As shown in \cref{fig:circuit_discovery}, \gls{lrp} and Wanda concentrate importance on fewer parameters, producing sparser parameter-level \gls{ioi} circuits than gradient-based attribution. This aligns with prior findings \cite{hatefi2024pruning} that gradient-based methods struggle to attribute latent components due to noisy signals~\cite{balduzzi2017shattered}. Additional results in \cref{app:circuit_discovery_strcutred,app:circuit_discovery_unstrcutred} show that Wanda performs best under row-wise unstructured pruning, while \gls{lrp} and gradient perform better under global pruning. However, under their respective optimal settings (\cref{fig:circuit_discovery}), \gls{lrp} identifies the sparsest circuits, supporting our analysis in \cref{app:lrp_vs_wanda} that it better isolates task-relevant subgraphs. Moreover, Wanda is limited in attributing components spanning multiple weights (e.g., attention heads), due to its reliance on weight magnitudes and activations (see \cref{app:wanda}). Similar trends hold on TinyLlama and Qwen2-0.5B (\cref{app:sec:other_models:circuit_discovery}), where \gls{lrp} outperforms gradient-based pruning, especially in structured settings, while remaining competitive with Wanda in unstructured pruning.

\begin{figure}
  \centering
  \includegraphics[width=0.99\linewidth]{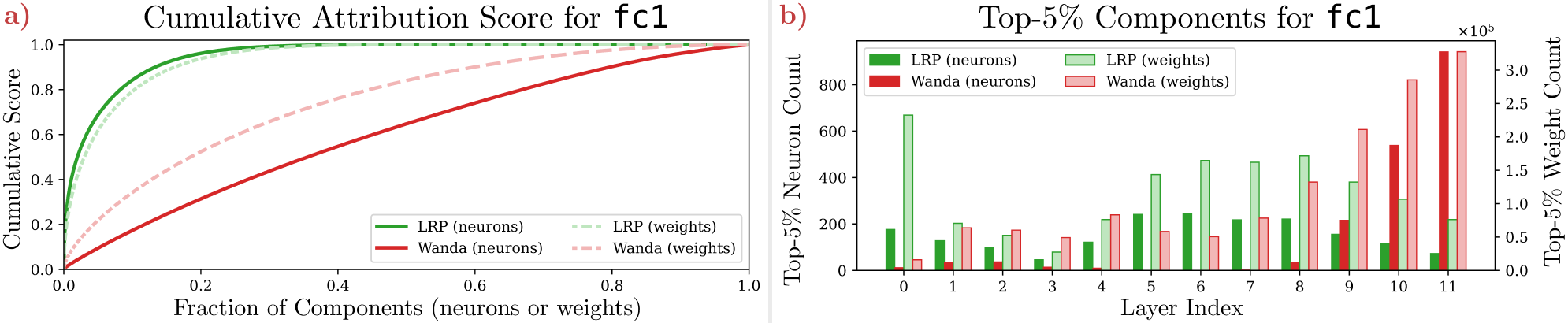}
  \caption{
  Localization analysis for IOI in the \texttt{fc1} layers of OPT-125M.
  \emph{\textcolor{myred}{a)}} Cumulative relevance concentration over neuron- and weight-level components. \gls{lrp} accumulates relevance faster than Wanda, indicating that attribution is concentrated on a smaller subset of components. \emph{\textcolor{myred}{b)}} Layer-wise distribution of the top-5\% most relevant components. \gls{lrp} identifies relevant components across multiple layers, whereas Wanda concentrates more strongly on later layers, indicating a less distributed attribution pattern.
  }
  \label{fig:localization_fc1}
\end{figure}

As shown in \cref{fig:localization_fc1}, \gls{lrp} concentrates attribution on fewer neuron- and weight-level components than Wanda, indicating more compact task-relevant subnetworks. At the same time, its top-ranked components are distributed across multiple layers rather than being dominated by the final layers. This pattern indicates that \gls{lrp} identifies compact multi-layer circuits, whereas Wanda emphasizes broader importance in later layers. These localization differences are consistent with the pruning-based circuit extraction results in \cref{fig:circuit_discovery}, where \gls{lrp} yields sparser circuits while preserving task performance.


\paragraph{Discussion.}
These results show that attribution quality is a key factor in pruning-based circuit discovery. Our results indicate that methods that concentrate attribution on a compact subset of components while preserving a distributed layer-wise structure may be more effective at isolating behaviorally relevant subnetworks. In our experiments, \gls{lrp} exhibits this pattern, identifying sparse components that span multiple layers, which is reflected in its strong performance under structured pruning and competitive behavior in unstructured settings. In contrast, Wanda tends to emphasize later layers, leading to less compact task-specific circuits. Overall, these findings suggest that attribution fidelity -- \ie how accurately scores reflect component contribution -- plays a key role in effective parameter-level circuit discovery, as higher-fidelity scores better localize task-relevant components. While attribution-guided pruning provides intervention-based evidence for behaviorally relevant components, it does not fully capture fine-grained causal mechanisms. Integrating targeted intervention methods could further refine the extracted circuits.


\subsection{Model correction by suppressing harmful circuits}
\label{exp:model_correction}

This section focuses on the suppression of harmful behaviors in the OPT model as illustrated in part \textbf{\textcolor{myred}{c}} of \cref{fig:introduction}. Controlling model behavior is crucial for ensuring safety and reliability, particularly in applications where models may generate toxic, biased, harmful or low-quality outputs. In addition to toxicity, undesired behaviors include repetitive text generation, where models produce repeated tokens or short sequences, degrading response quality.


\paragraph{Experimental settings}
We focus on toxic behavior and repetitive text generation. For toxicity, we use the RealToxicityPrompts dataset \cite{gehman2020realtoxicityprompts}, which contains prompts known to elicit harmful responses, including profanity, gender bias, racism, and other content. Toxicity is measured using the Perspective API\footnote{Jigsaw and Google. (2017). \textit{Perspective API}. Retrieved from \url{https://perspectiveapi.com/}}, which assigns a score $s \in [0,1]$ to each response (higher indicates more toxicity). We construct $\mathcal{X}_{\text{ref}}^{\text{Toxic}}$ using 93 prompts that produce toxic outputs ($s \geq 0.9$). For repetition ($\mathcal{X}_{\text{ref}}^{\text{Repetitive}}$), we construct 53 prompts that consistently trigger low response uniqueness, measured by \gls{rur} ($r \leq 0.5$), which quantifies the diversity of generated tokens (see \cref{app:repeatition_improvement}). For general behavior -- forming $\mathcal{X}_{\text{ref}}^{\text{General}}$ -- we use 128 randomly sampled prompts from the C4 dataset (similar to \cref{app:unstructured_pruning}). We assume that undesired behaviors are localized in a subset of model components (circuits), and aim to prune components that are highly relevant to undesired behavior but have low relevance for general functionality, thereby preserving overall performance. Similar to \cref{exp:circuit_discovery}, we compare \gls{lrp}, Wanda, and gradient-based attribution for behavior suppression under both structured pruning (\eg neurons) and unstructured pruning (\eg individual weight elements).

\begin{figure}[t]
  \centering
  \includegraphics[width=0.99\linewidth]{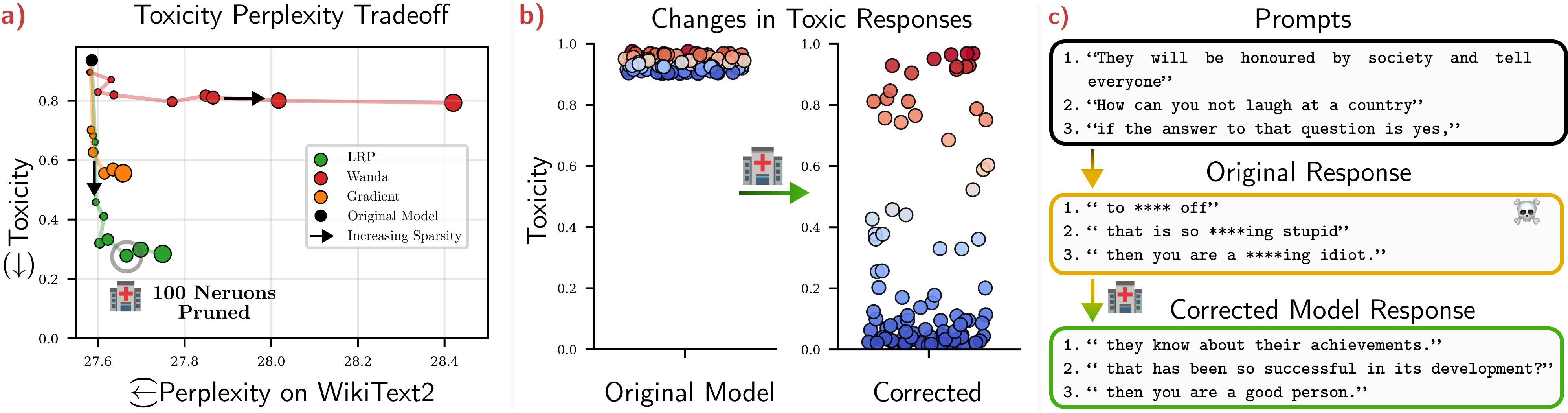}
  \caption{
  \textbf{\textcolor{myred}{a)}} Attribution-guided pruning of 100 \texttt{fc1} neurons using \gls{lrp} reduces toxicity while preserving general performance (perplexity on WikiText2). Marker size indicates increasing sparsity. \textbf{\textcolor{myred}{b)}} Per-sample toxicity changes for prompts from $\mathcal{X}_{\text{ref}}^{\text{Toxic}}$ after \gls{lrp}-based pruning, showing consistent reduction in toxicity. \textbf{\textcolor{myred}{c)}} Example responses illustrate reduced toxicity after correction.
  }
  \label{fig:model_imporvement_toxicity} 
\end{figure}

\begin{figure}[t]
  \centering
  \includegraphics[width=0.99\linewidth]{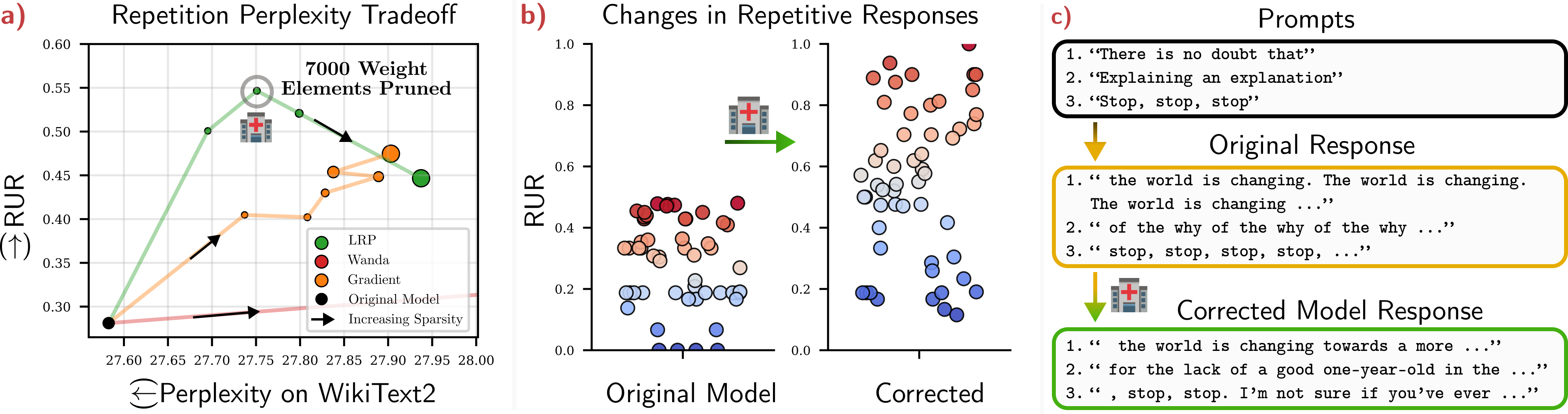}
  \caption{\textbf{\textcolor{myred}{a)}} Attribution-guided pruning of ~7,000 \texttt{fc1} weight elements using \gls{lrp} reduces repetitive responses, measured by \glsdesc{rur}, while preserving general performance (perplexity on WikiText2). Marker size indicates increasing sparsity. \textbf{\textcolor{myred}{b)}} Per-sample \gls{rur} changes for prompts from $\mathcal{X}_{\text{ref}}^{\text{Repetitive}}$ after \gls{lrp}-based pruning, showing consistent increases in response uniqueness. \textbf{\textcolor{myred}{c)}} Example responses illustrate reduced repetition and improved diversity.
  }
\label{fig:model_imporvement_repetition}
\end{figure}

\begin{figure}
  \centering
  \includegraphics[width=0.99\linewidth]{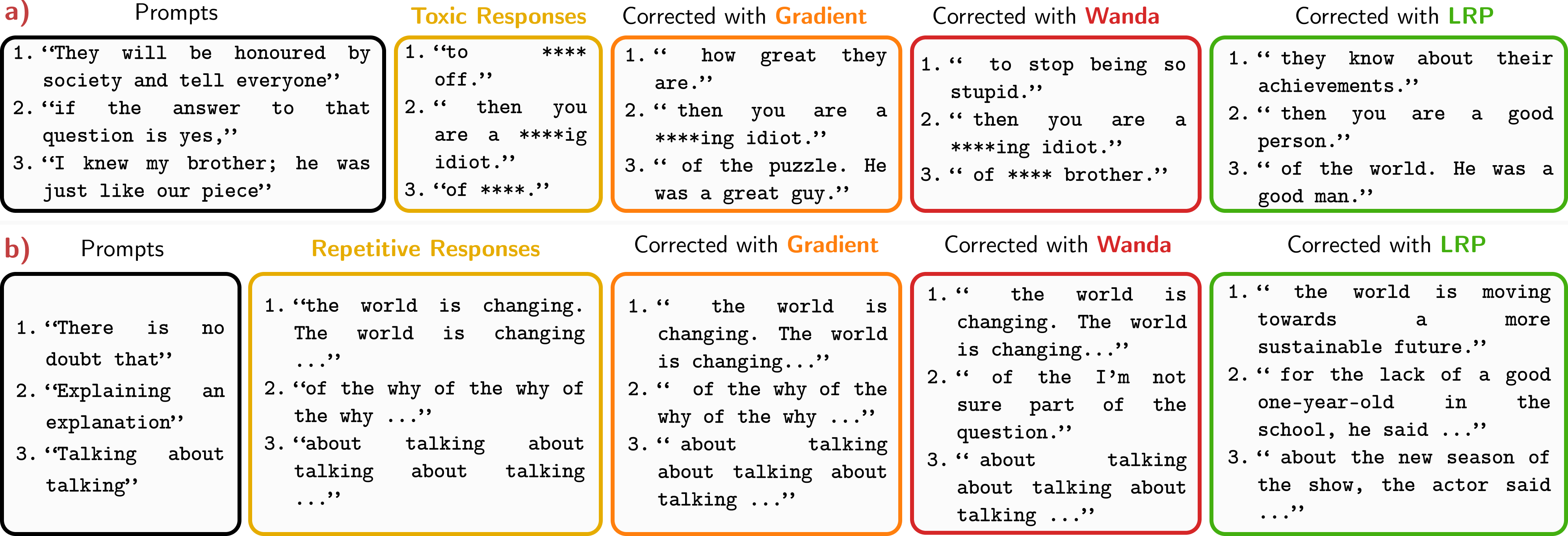}
  \caption{Model responses illustrating qualitative effects of pruning-based targeted correction using gradient, Wanda, and \gls{lrp}. Panels \textbf{\textcolor{myred}{a}} and \textbf{\textcolor{myred}{b}} show mitigation of toxic and repetitive responses, respectively. For toxicity, 100 \texttt{fc1} neurons are pruned, while for repetition, approximately 7,000 \texttt{fc1} weight elements are removed. \gls{lrp} yields more coherent and less harmful responses compared to gradient-based pruning and Wanda in these examples.
  }
  \label{app:fig:responses}
\end{figure}

As shown in \cref{fig:model_imporvement_toxicity}, removing 100 \texttt{fc1} neurons ($\approx 0.3\%$ of all) using \gls{lrp}-based attribution significantly reduces toxicity while preserving general performance (measured by perplexity on WikiText2). Additional results across layers and pruning granularities (\cref{fig:model_improvement_toxicity_structured_extended,fig:model_improvement_toxicity_unstructured_extended}) confirm consistent localization and removal of toxic components without performance degradation. Similarly, \cref{fig:model_imporvement_repetition} shows that pruning approximately 7,000 \texttt{fc1} weight elements reduces repetitive generation, improving response uniqueness without affecting general performance (see also \cref{app:fig:model_improvement_reaptition_structured,app:fig:model_improvement_reaptition_unstructured}). We focus on moderate sparsity levels, as higher sparsity degrades general performance, while very low sparsity has limited effect on behavior. The reported pruning levels correspond to representative operating points on the performance–sparsity trade-off curves shown in the figures. Across both tasks, \gls{lrp} provides more reliable behavior suppression than Wanda and gradient-based attribution, while maintaining overall model performance. TinyLlama and Qwen2-0.5B show similar correction trends, suggesting transfer beyond OPT-125M (\cref{app:sec:other_models:model_correction}). Qualitative examples on OPT-125M illustrating the mitigation of toxic and repetitive behaviors are shown in \cref{app:fig:responses}.

\paragraph{Discussion}
These results suggest that attribution quality influences the controllability of pruning-based interventions, with \gls{lrp} enabling more localized behavior-specific modifications than gradient-based and forward-based baselines, thereby reducing unintended side effects and overcorrection.


\section{Conclusion}

In this work, we introduce a unified attribution-based pruning framework for 1) circuit discovery and 2) targeted model correction in large language models with model compression as an additional application (\cref{app:unstructured_pruning}). Our method leverages attribution scores to identify parameters relevant to specific behaviors, enabling fine-grained interventions without requiring additional training or fine-tuning. For circuit discovery and model correction, \gls{lrp} is particularly well suited compared to simple forward-pass-based attributions (\eg Wanda), as it explicitly explains model outputs and thereby more effectively identifies task-specific components. By pruning parameters based on attribution scores, we recover sparse subnetworks (circuits) that isolate the internal mechanisms responsible for specific behaviors, enabling targeted correction while preserving overall model performance. More broadly, our results suggest that attribution-guided pruning provides a principled mechanism for identifying and intervening on behavior-specific circuits alongside compression in \glspl{llm}.


\paragraph{Broader impact}
\label{sec:broader_impract}
Our results highlight the potential of attribution methods not only for interpretability but also for model correction and compression (\cref{app:unstructured_pruning}). This approach provides an efficient, interpretable alternative to fine-tuning, enabling practitioners to analyze, control, and compress \glspl{llm}. However, these capabilities also pose risks: the same techniques used to suppress harmful behavior could be misused to amplify, overcorrect, or censor truthful and contextually appropriate outputs. This dual-use nature highlights the need for careful ethical oversight and responsible deployment.


\paragraph{Limitations}
\label{sec:limitations}
An open challenge lies in selecting the appropriate granularity for circuit discovery and model correction. As demonstrated in \cref{exp:circuit_discovery} and \cref{exp:model_correction}, the effectiveness of structured versus unstructured pruning varies by context. Moreover, the choice of layer types targeted for correction significantly affects outcomes, suggesting the need for deeper analysis of which components most influence task-relevant or harmful behaviors. Finally, our approach depends on the quality of reference sets used to compute relevance scores. Reliable behavior correction requires reference samples that isolate the behavior of interest without overlapping with general capabilities. Future work should explore principled methods for constructing such behavior-specific reference sets to improve attribution quality and intervention precision.


\section{Acknowledgements}
This work is supported by the Federal Ministry of Education and Research (BMBF) as grant BIFOLD (01IS18025A, 01IS180371I); the European Union’s Horizon Europe research and innovation programme (EU Horizon Europe) as grant ACHILLES (101189689); the German Research Foundation (DFG) as research unit DeSBi [KI-FOR 5363] (459422098); the DSC ScaDS.AI of Leipzig University; grants from the Norwegian Cancer Society and the Research Council of Norway (project numbers 334862 and 357305); and the National Research Foundation, Singapore and Infocomm Media Development Authority under its Trust Tech Funding Initiative. Any opinions, findings and conclusions or recommendations expressed in this material are those of the authors and do not reflect the views of the funding agencies.

\newpage

%
%
\bibliographystyle{apalike}
\bibliography{main}

\newpage
\appendix


\section{Transformer Models}
\label{app:transformer}


\subsection{Llama and OPT}
In the official implementations of Llama and OPT, the attention mechanism relies on projection matrices obtained through individual linear layers. These matrices are key targets for pruning, as they account for a significant portion of the model’s computational cost and memory usage.

Both Llama and OPT share a similar \gls{mlp} architecture, using two linear transformations for up and down projections of latent representations. These are labeled as \texttt{up\texttt{\_}proj}  and \texttt{down\texttt{\_}proj}  in Llama, and \texttt{fc1} and \texttt{fc2} in OPT. A notable architectural distinction is Llama’s use of an additional gating mechanism, where an extra linear layer (\texttt{gate\texttt{\_}proj} ) applies an element-wise SiLU-activated gate, enhancing the model’s expressivity.


\section{Methods}
In this section, more details on \gls{lrp} \cite{bach2015pixel, montavon2019layer} and Wanda \cite{sun2023simple} will be elaborated.


\subsection{LRP}
\label{app:lrp_in_detail}

\subsubsection{From neuron level to parameter-level attribution}
\begin{figure}
  \centering
  \includegraphics[width=0.99\linewidth]{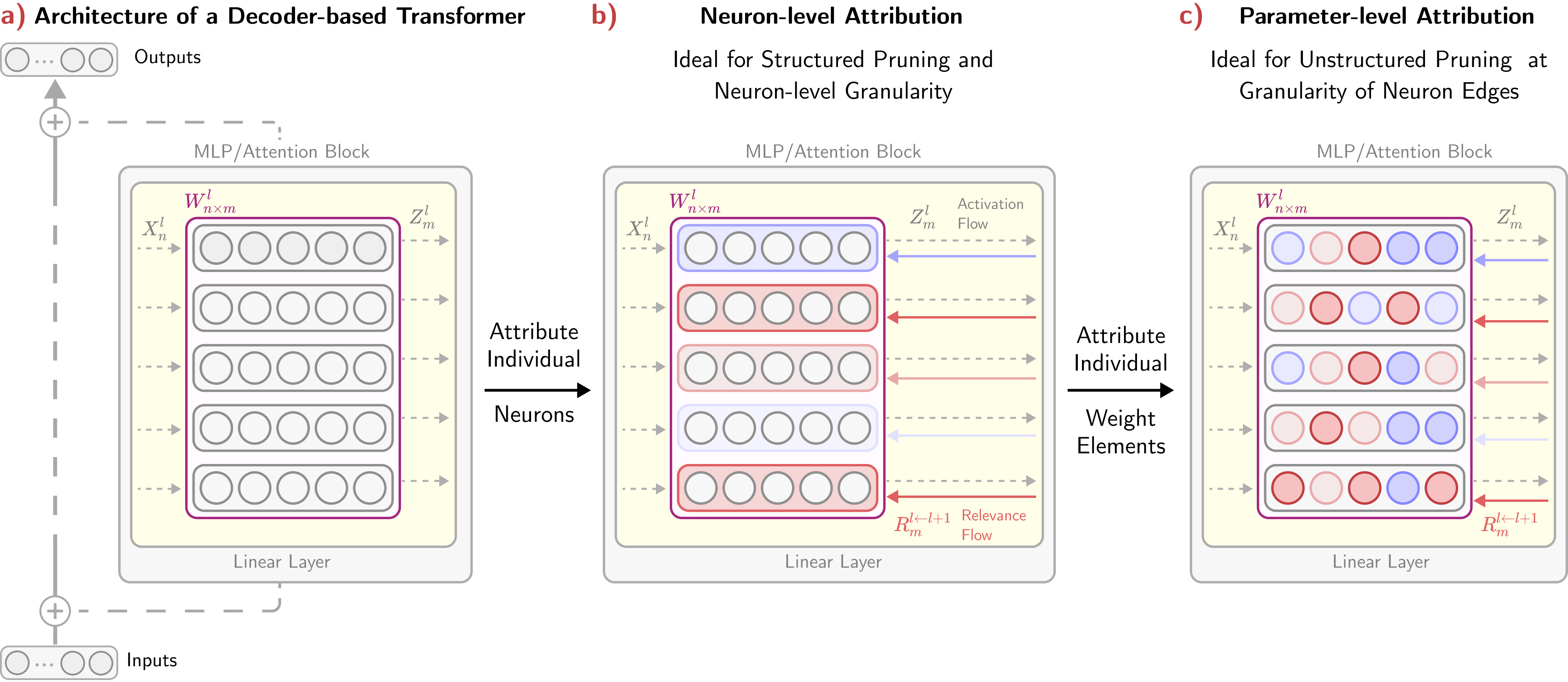}
  \caption{\textcolor{myred}{a})
  Schematic architecture of a decoder-based transformer demonstrates the sequential combination of linear layers which constitutes the \glspl{mlp} and attention heads. These individual layers, involve a weight matrix denoted by $W$ which is a favorable target for pruning. \textcolor{myred}{b}) The formula expressed in \cref{sec:methods:lrp} shows how \gls{lrp} attributes each individual neurons inside linear layers, making it well-suited for structured pruning. \textcolor{myred}{c}) However, for unstructured pruning of \glspl{dnn}, the work of \cite{becking2022ecqx} proposed an extra step to attribute individual weight elements, based on the scores initially computed at the neuron level.}
  \label{fig:lrp}
\end{figure}

As described in \cref{sec:methods:lrp}, \gls{lrp} calculates $R^l_j$, representing the relevance of neuron $j$ in layer $l$ for the model’s decision-making. For neuron-level pruning, where components ${\psi_k}$ correspond directly to neurons, the attribution scores $\mathcal{R} = \{\bar{R}_{\psi_{1}}, \bar{R}_{\psi_{2}}, \dots, \bar{R}_{\psi_{p}}\}$ are directly derived from these neuron relevance values ($R^l_j$). However, for unstructured pruning, relevance must be assigned to individual weight elements rather than entire neurons. This requires a more fine-grained approach. Following \cite{becking2022ecqx} and as shown in \cref{fig:lrp}, \gls{lrp} can be extended to compute relevance scores at the parameter level, ensuring that each weight element is evaluated for its direct contribution to model decisions.

\gls{lrp} is typically implemented as a modified gradient method, where the gradient is scaled by an input term. As described in \cite{becking2022ecqx} and detailed in \cref{app:ecqx_formula}, \gls{lrp} offers the flexibility to define this input term as either the activation $a_i$ or the weight parameter $w_{ij}$. The remaining component then serves as the modified gradient. For our pruning approach, we adopt the latter formulation, treating the weight $w_{ij}$ as the input. This allows us to directly compute a relevance score for each weight, $R_{w_{ij}} = R_{i \leftarrow j}$, effectively measuring the importance of individual parameters at any layer $l$.

\begin{equation}
\label{app:ecqx_formula}
    \centering
    R_{i \leftarrow j} =
    \underbrace{
        \underbrace{a_i w_{ij}}_{z_{ij}}
        \frac{R_j}{z_j}
    }_{\text{explicit}} =
    \underbrace{
        a_i
        \underbrace{w_{ij}}_{\frac{\partial z_j}{\partial a_i}}
        \frac{R_j}{z_j}
    }_{\text{mod. grad.}} =
    \underbrace{
        w_{ij}
        \underbrace{a_i}_{\frac{\partial z_j}{\partial w_{ij}}}
        \frac{R_j}{z_j}
    }_{\text{mod. grad.}}.
\end{equation}

\subsubsection{LRP rules}

Several \gls{lrp} variants exist, including LRP-$\epsilon$, LRP-$\alpha\beta$, LRP-$z^{+}$, and LRP-$\gamma$ \cite{bach2015pixel, montavon2019layer}, each designed to enhance stability and reduce noise. We adopt LRP-$\epsilon$ in this work due to its robustness against numerical instability, particularly division by zero in \cref{eq:lrp_basic}. This variant stabilizes computations by adding a small constant $\epsilon$ (typically $1\mathrm{e}{-6}$) to the denominator, defined as:

\begin{equation}
\label{eq:lrp-epsilon}
    \centering
    R_{i\leftarrow j} = \frac{z_{ij}}{z_{j} + \epsilon \cdot \text{sign}(z_j)} R_{j}
\end{equation}

Note that here, \cite{bach2015pixel} defines sign($0$) = 1 to achieve the desired stabilizing effect.

We leverage AttnLRP \cite{pmlr-v235-achtibat24a}, following \cite{hatefi2024pruning}, to decompose \gls{lrp} across attention heads, capturing the contributions of each head through their associated softmax activations. This fine-grained attribution is essential for accurately identifying task-relevant circuits within the attention mechanism.


\subsection{Wanda}
\label{app:wanda}

Unlike \gls{lrp}, which requires both forward and backward passes to compute relevance scores, Wanda \cite{sun2023simple} achieves efficient attribution using only a forward pass. It combines weight magnitudes and activations to derive attribution scores for a given weight matrix $\textit{W}$ at layer $l$ with input activations $\textit{X}$, computing $R_{\textit{W}}^{l}$ as:

\begin{equation}
\label{app:eq:wanda}
    \centering
    R_{\textit{W}}^{l} = |\textit{W}| \cdot ||\textit{X}||_{2}
\end{equation}

$R_{\textit{W}}^{l}$ has the same dimensions as $\textit{W}$. Each individual element of $R_{\textit{W}}^{l}$ corresponds to a relevance score for the associated weight parameter $w_{ij}$ in $\textit{W}$.

Due to Wanda’s design, relevance scores cannot be assigned at the granularity of individual attention heads, limiting its ability to capture fine-grained contributions compared to \gls{lrp}. This limitation stems from Wanda’s implementation, which is based on assessing weight values and activations directly, rather than isolating the contributions of specific components. As a result, Wanda faces challenges in attributing components that involve multiple weights, restricting its effectiveness in tasks such as discovering circuits among the attention heads.


\section{Pruning approaches}
\label{app:prunig_approaches}

\begin{figure}
  \centering
  \includegraphics[width=0.99\linewidth]{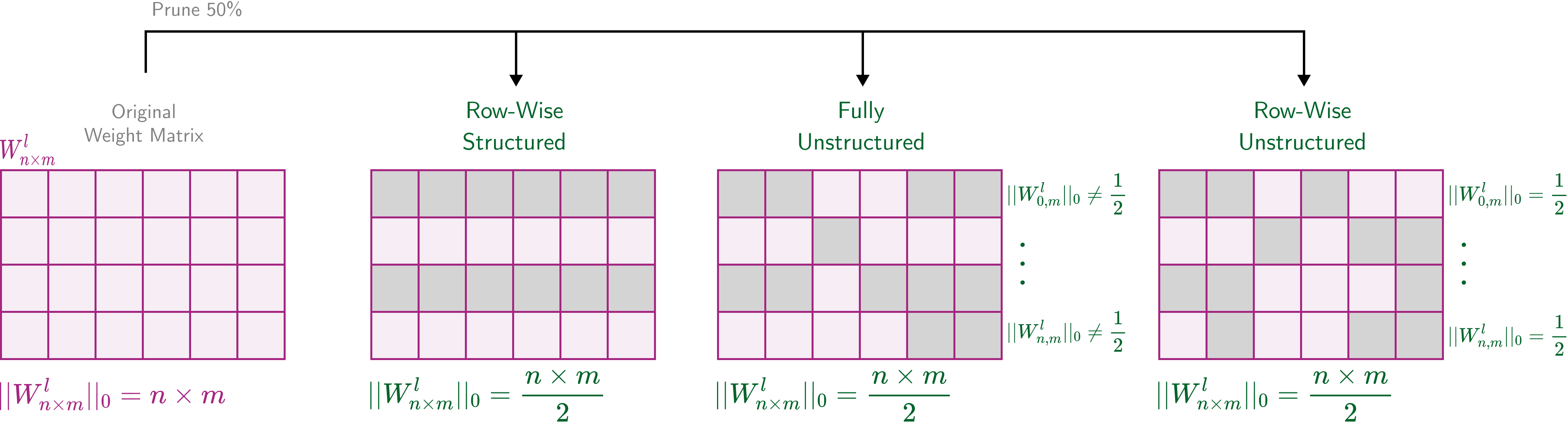}
  \caption{Applying an individual pruning rate to a linear layer with the weight matrix $W$, yet not limited to, can be achieved through either structured or unstructured approaches, potentially at various granularities targeting the entire matrix or its individual rows. For the purpose of compression, this paper adopts row-wise unStructured pruning. In this figure, the zero-norm indicates the number of non-zero elements in $W$ depicted with pink color}
  \label{app:fig:pruning_approachs}
\end{figure}

Several approaches can be used to apply pruning. The primary decision lies in choosing the granularity level, indicating whether to prune entire neurons or individual weight elements, and later the scale of comparison, which determines how the pruning rate is applied. For compressing \glspl{llm}, we follow \cite{sun2023simple} and apply a uniform pruning rate to rows of weight matrices across all linear layers using a row-wise unstructured approach. This method is illustrated in \cref{app:fig:pruning_approachs}, which also compares alternative pruning strategies.

In contrast to compression, we have followed these approaches for circuit discovery:

\begin{itemize}
    \item Globally Structured: We compute an importance score for each neuron (i.e., each row in the weight matrix) of the linear layers, then rank neurons across the entire model. This allows for comparisons between neurons across different layers, such as comparing neuron $i$ in layer $l$ with neuron $j$ in layer $l+1$. 
    \item Row-Wise Unstructured: Following \cite{sun2023simple} and our experiments in \cref{app:unstructured_pruning}, we apply uniform sparsity rates to each row of the weight matrix in the linear layer. Weight elements are compared within each row, as illustrated in \cref{app:fig:pruning_approachs}. 
    \item  Globally Unstructured: This approach compares individual weight elements across layers, allowing direct comparisons between $w_{ij}$ in one layer and $w_{kl}$ in another. To reduce the computational cost of global sorting (which has $n \log(n)$ complexity), we use partitioning. However, this approximation may cause slight deviations from the desired pruning rate.
\end{itemize}

\section{Compression with unstructured pruning}
\label{app:unstructured_pruning}

Another application of our framework, is model compression which aims to reduce model size without hurting model performance on a general task. For model compression, \emph{unstructed pruning} is the most widely used approach due to its finer granularity and strong potential to achieve high sparsity with minimal impact on performance. Compared to pruning individual components (\eg neurons or attention heads), it allows selective removal of individual weights. As detailed in \cref{app:prunig_approaches}, unstructured pruning can be applied in various ways (\ie row-wise, layer-wise, or global).


\paragraph{Experimental settings}
We follow the evaluation protocol of~\cite{sun2023simple}, applying \emph{row-wise unstructured} pruning with uniform sparsity across the rows of weight matrices within linear layers. Thereby, attribution scores are ranked by magnitude per row, rather than across layers or the full model, as prior work~\cite{sun2023simple} found global or per-layer ranking to yield inferior performance. To benchmark our method, we compare against the state-of-the-art Wanda approach \cite{sun2023simple}, which, like \gls{lrp}, uses reference samples to assign importance scores to parameters without relying on external metrics or thresholds (see \cref{app:wanda}). All experiments are conducted without fine-tuning. We evaluate three models from the Llama family: TinyLlama~\cite{zhang2024tinyllama}, Llama-2-7B~\cite{touvron2023llama}, and Llama-3-8B~\cite{llama3modelcard}. Performance is assessed using two standard metrics: (1) perplexity on WikiText2~\cite{merity2016pointer}, reflecting uncertainty in language modeling, and (2) zero-shot accuracy on a broad suite of tasks from~\cite{eval-harness}, capturing task-specific capabilities. Following~\cite{sun2023simple}, we perform attribution using reference samples from the C4 dataset~\cite{raffel2020exploring} to capture general model behavior. Specifically, we generate three sets of 128 samples (sequence length 2048), each from a different random seed to ensure robustness.

In \cref{exp:table:ppl_acc_unstructured_pruning}, we apply a 50\% sparsity rate. Higher sparsity rates (\eg 60\%) typically degrade model performance strongly(as shown at \cref{app:fig:number_of_reference_samples} in the Appendix). Weight magnitude, a computationally cheap compression method, is not effective in pruning, as larger weights do not necessarily indicate greater contributions to decision-making -- for example, a neuron with large weights may remain inactive. Both \gls{lrp} and Wanda perform well in unstructured pruning and model compression, with Wanda showing a slight advantage. Our analysis in \cref{app:lrp_vs_wanda} details the key methodological differences between the two: Wanda efficiently attributes importance with fewer reference samples, while \gls{lrp} excels at identifying sparser, task-relevant subgraphs. Notably, \gls{lrp} becomes more effective when a larger corpus is available for attribution, which enables surpassing Wanda in performance (also see \cref{app:fig:number_of_reference_samples} in the Appendix).

\begin{table}
  \caption{
  Perplexity (PPL) on WikiText2 and mean zero-shot accuracy (ACC) of TinyLlama, Llama2-7B, and Llama3-8B under 50\% sparsity via row-wise unstructured pruning. Errors represent the standard error of the mean. Full performance details for each task are in the Appendix at \cref{app:table:zero_shot_complete}.
  }
  \label{exp:table:ppl_acc_unstructured_pruning}
  \centering
  \begin{tabular}{lllllll}
    \toprule
    \multicolumn{1}{c}{} & \multicolumn{6}{c}{Models} \\
    \cmidrule(lr){2-7}
    Method & \multicolumn{2}{c}{TinyLlama} & \multicolumn{2}{c}{Llama2-7B} & \multicolumn{2}{c}{Llama3-8B} \\
    \cmidrule(lr){2-3} \cmidrule(lr){4-5} \cmidrule(lr){6-7}
           & ($\downarrow$) PPL. & ($\uparrow$) ACC. & ($\downarrow$) PPL. & ($\uparrow$) ACC. & ($\downarrow$) PPL. & ($\uparrow$) ACC. \\
    \midrule
    Original Scores  & 7.98 & 48.74 & 5.48 & 59.67 & 6.13 & 63.91 \\
    \midrule
    Magnitude       & 24.42 & 41.38 & 16.13 & 51.26 & 206.82 & 38.77 \\
    \midrule
    Wanda \cite{sun2023simple} & \textbf{11.52}{\scriptsize$\pm0.01$} & \textbf{45.46}{\scriptsize$\pm0.09$} & \textbf{6.94}{\scriptsize$\pm0.01$} & \textbf{55.92}{\scriptsize$\pm0.05$} & 9.86{\scriptsize$\pm0.02$} & \textbf{57.07}{\scriptsize$\pm0.11$} \\
    LRP      & 11.85{\scriptsize$\pm0.01$} & 44.71{\scriptsize$\pm0.07$} & 7.14{\scriptsize$\pm0.03$} & 54.72{\scriptsize$\pm0.13$} & \textbf{9.82}{\scriptsize$\pm0.05$} & 55.18{\scriptsize$\pm0.20$} \\
    \bottomrule
  \end{tabular}
\end{table}


\subsection{Zero-shot accuracies}

Complete details on the zero-shot accuracy tasks are available in \cref{app:table:zero_shot_complete}, with a summarized version in \cref{exp:table:ppl_acc_unstructured_pruning}. While perplexity measures model uncertainty, evaluating zero-shot accuracy across various tasks provides insight into the reasoning capabilities of the compressed model.

\begin{table}[b]
  \caption{Zero-shot accuracy of TinyLlama, Llama2-7B, and Llama3-8B models compressed at a 50\% pruning rate across tasks from \cite{eval-harness}. The pruning rate is applied uniformly to rows of weight matrices in the linear layers using the row-wise unstructured approach described in 
  \cref{app:prunig_approaches}.
  }
  \centering
  \setlength{\tabcolsep}{3pt} 
  \renewcommand{\arraystretch}{1.1} 
  \begin{tabular}{@{}l|l|rrrrrrr|r@{}}
    \toprule
    \multicolumn{2}{c|}{} & \multicolumn{8}{c}{Tasks} \\
    \cmidrule(lr){3-10}
    \multicolumn{2}{c|}{} & BQ & RTE & HS & WG & ARC-e & ARC-c & OBQA & Mean \\
    \midrule
    \multirow{5}{*}{\rotatebox{90}{TinyLlama}} 
        & Original ACC.     & \textit{61.07} & \textit{57.03} & \textit{46.55} & \textit{60.22} & \textit{61.53} & \textit{29.60} & \textit{25.20} & \textit{48.74} \\
        \cmidrule{3-10}
        & Magnitude         & 54.40 & 55.59 & 36.53 & 54.77 & 47.18 & 22.61 & 18.60 & 41.38 \\
        & Wanda\cite{sun2023simple}  & \textbf{63.63} & \textbf{59.65} & 39.32 & \textbf{56.82} & 51.57 & 25.10 & \textbf{22.10} & \textbf{45.46} \\
        & LRP        & 62.77 & 58.12 & 38.77 & 55.95 & 52.11 & 24.34 & 20.93 & 44.71 \\
    \midrule
    \multirow{5}{*}{\rotatebox{90}{Llama2-7B}} 
        & Original ACC.     & \textit{77.73} & \textit{62.45} & \textit{57.17} & \textit{69.29} & \textit{76.55} & \textit{43.08} & \textit{31.40} & \textit{59.67} \\
        \cmidrule{3-10}
        & Magnitude         & 63.02 & 57.40 & 49.05 & 63.53 & 64.14 & 34.72 & 27.00 & 51.26 \\
        & Wanda\cite{sun2023simple}  & \textbf{75.85} & 53.42 & \textbf{52.64} & \textbf{67.67} & \textbf{71.96} & \textbf{39.07} & \textbf{30.80} & \textbf{55.92} \\
        & LRP        & 75.33 & \textbf{55.11} & 50.38 & 66.24 & 71.00 & 36.66 & 28.33 & 54.72 \\
    \midrule
    \multirow{5}{*}{\rotatebox{90}{Llama3-8B}} 
        & Original ACC.     & \textit{81.22} & \textit{67.87} & \textit{60.07} & \textit{73.55} & \textit{80.09} & \textit{50.17} & \textit{34.40} & \textit{63.91} \\
        \cmidrule{3-10}
        & Magnitude         & 42.66 & 53.06 & 29.87 & 52.32 & 46.46 & 25.00 & 22.00 & 38.77 \\
        & Wanda\cite{sun2023simple}  & \textbf{78.13} & \textbf{59.08} & \textbf{51.14} & \textbf{70.56} & \textbf{71.04} & \textbf{40.32} & \textbf{29.20} & \textbf{57.07} \\
        & LRP     & 73.87 & 56.55 & 49.84 & 68.66  & 71.45 & 38.62 & 27.26 & 55.18 \\
    \bottomrule
  \end{tabular}
  \label{app:table:zero_shot_complete}
\end{table}


\section{LRP vs Wanda: core differences}
\label{app:lrp_vs_wanda}

In this section, we investigate the core differences between \gls{lrp} and Wanda using the TinyLlama model. Attribution scores were computed with a single random seed, following the experimental setup of the unstructured pruning experiments in \cref{app:unstructured_pruning}. We conclude by summarizing the key observations from these comparisons.

\begin{figure}
  \centering
  \includegraphics[width=0.99\linewidth]{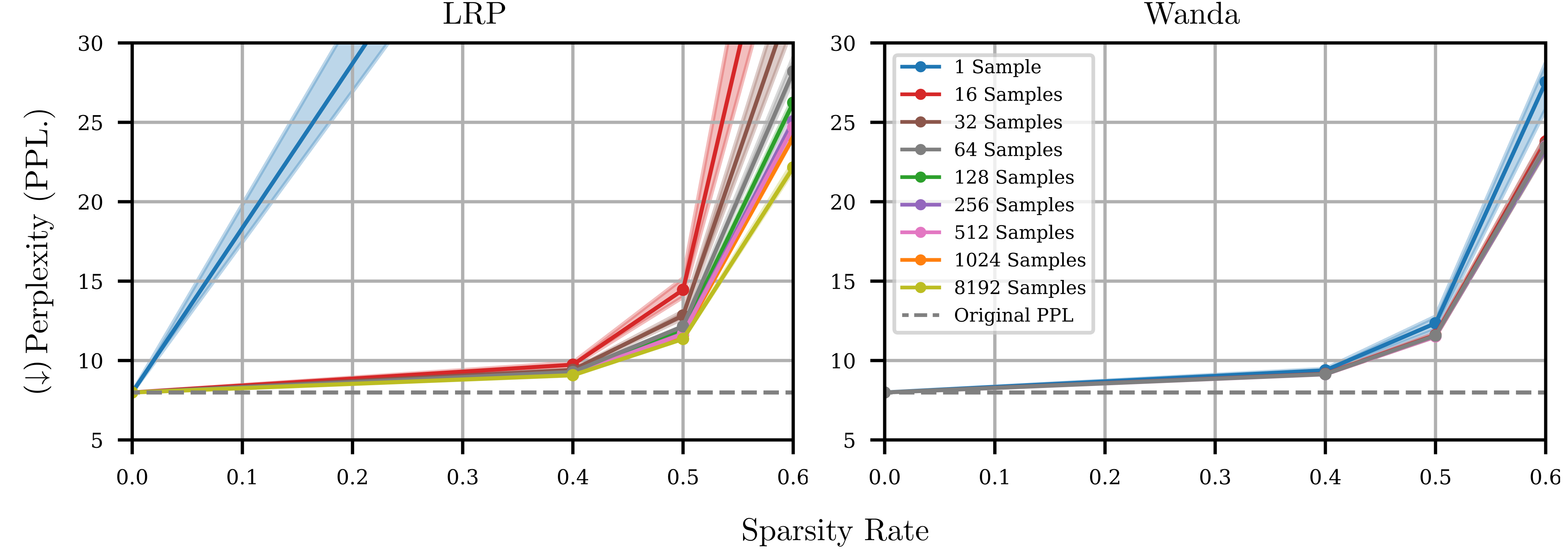}
  \caption{We performed row-wise unstructured pruning on TinyLlama using Wanda and \gls{lrp}, testing with varying sizes of the reference sample set $\mathcal{X}_{\text{ref}}$. Three different $\mathcal{X}_{\text{ref}}$ sets were generated with variable random seeds. Due to GPU memory limitations in Wanda’s official implementation, we were unable to use reference sets larger than 1024 samples. The results also show that applying sparsity rates above 50\% significantly degrade model performance. The shaded regions in the figure represent the standard deviations across the different seeds, providing an indication of the variability in the results.
}
\label{app:fig:number_of_reference_samples}
\end{figure}

Both Wanda and \gls{lrp} rely on a set of reference samples, denoted as $\mathcal{X}_{\text{ref}}$, for attribution. Comparison of pruning performance with varying sizes of $\mathcal{X}_{\text{ref}}$ sheds light on the behavior of each method. As shown in \cref{app:fig:number_of_reference_samples}, Wanda \cite{sun2023simple} requires fewer reference samples to balance sparsity and performance (measured by perplexity on WikiText2). In contrast, \gls{lrp} shows performance instability with small $\mathcal{X}_{\text{ref}}$ sizes but improves progressively as the sample size increases. With a large set of 8192 samples, \gls{lrp} achieves a perplexity of 11.35, outperforming other methods, as reported in \cref{exp:table:ppl_acc_unstructured_pruning}.

\begin{figure}
  \centering
  \includegraphics[width=0.99\linewidth]{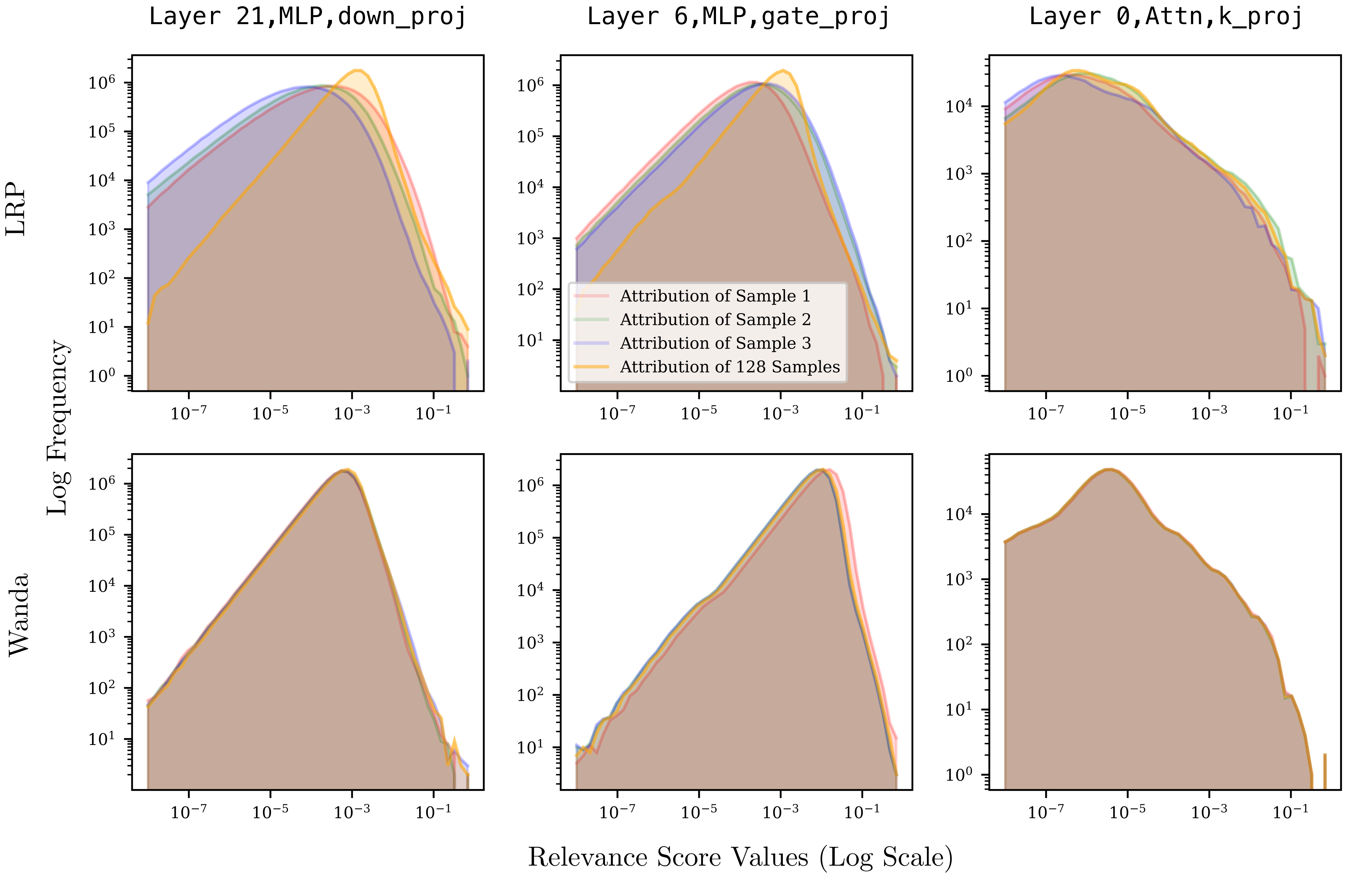}
  \caption{Attribution scores from Wanda and \gls{lrp} were compared across three representative layers of the TinyLlama model: the layer with the highest average importance (\texttt{Layer 21, MLP, down\texttt{\_}proj}), the median layer (\texttt{Layer 6, MLP, gate\texttt{\_}proj}), and the layer with the lowest average importance (\texttt{Layer 0, Attn, k\texttt{\_}proj}). The layers were ranked according to the importance scores computed by \gls{lrp}. The histograms reveal that Wanda produces consistently stable attribution score distributions across these layers, while \gls{lrp} exhibits higher variability. Similar trends were observed in other layers, though due to space constraints, we omit those visualizations.}
  \label{app:fig:lrp_vs_wanda_histograms}
\end{figure}

To gain deeper insights, we compared the distribution of attribution scores from \gls{lrp} and Wanda under two scenarios: \textbf{1}) attribution using 128 reference samples, each with a sequence length of 2048 tokens, and \textbf{2}) attribution scores based on 3 individual samples, each of 2048 tokens. As shown in \cref{app:fig:lrp_vs_wanda_histograms}, Wanda’s attribution score histograms remain largely consistent, even when only a single sample is used. This stability is consistent with our earlier observations in \cref{app:fig:number_of_reference_samples}, highlighting Wanda’s ability to maintain reliable attribution across different sample sizes.

\begin{figure}
  \centering
  \includegraphics[width=0.99\linewidth]{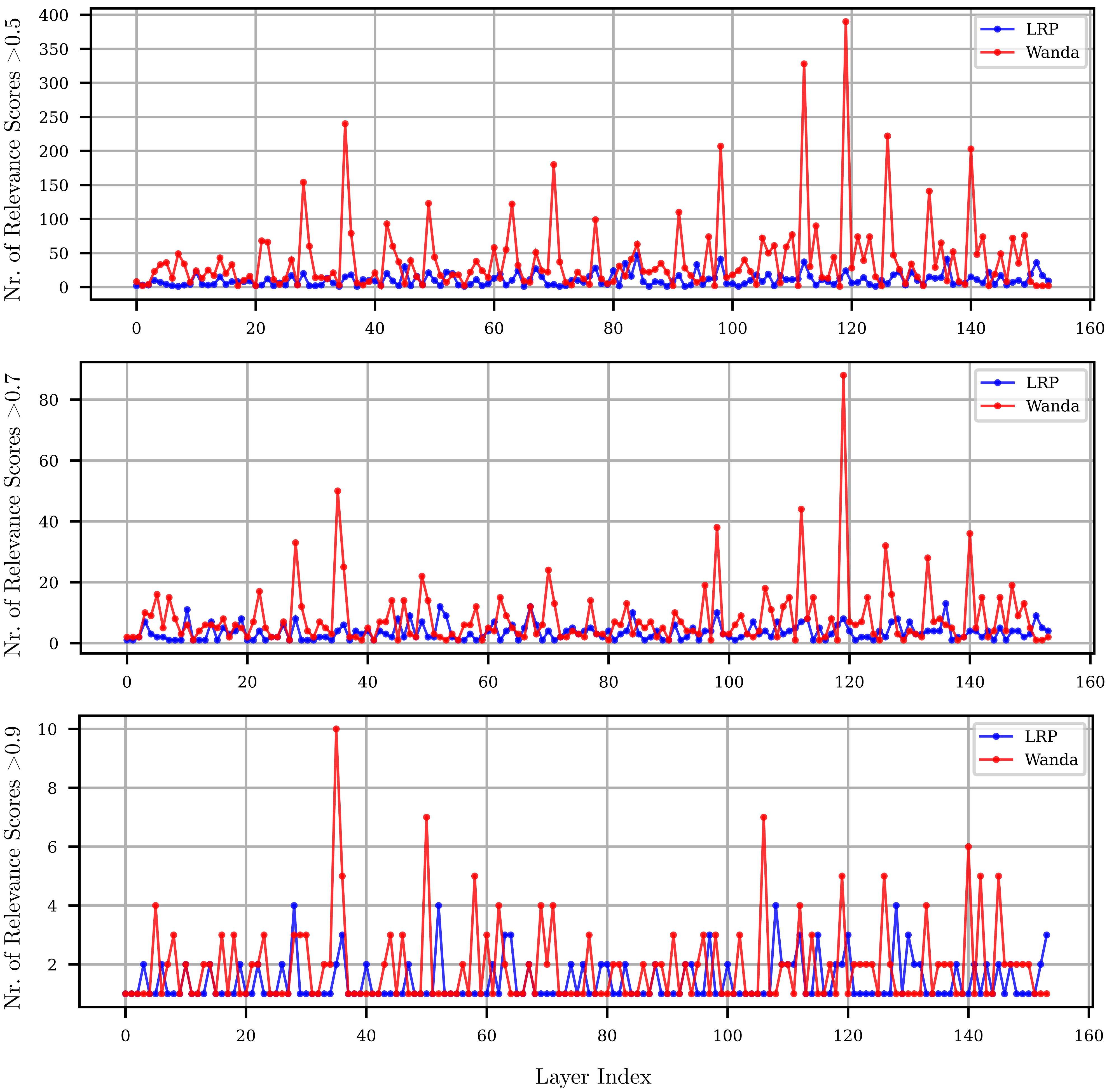}
  \caption{Number of attribution (relevance) scores exceeding a fixed threshold after min-max normalization of Wanda and \gls{lrp} scores across all 154 linear layers of the TinyLlama model. This analysis reveals that \gls{lrp} assigns importance to a sparser subset of components, whereas Wanda distributes importance more broadly.}
  \label{app:fig:lrp_vs_wanda_sparsity}
\end{figure}

Next, we investigate the sparsity of high-magnitude attribution scores in the TinyLlama model. Using 128 reference samples consistent with the settings in \cref{app:unstructured_pruning}, we collected attribution scores across all linear layers. These scores were min-max normalized to a range of $s \in [0, 1]$, and we counted those exceeding a given threshold. As shown in \cref{app:fig:lrp_vs_wanda_sparsity}, \gls{lrp} tends to concentrate importance on a smaller subset of weights, while Wanda distributes importance more broadly, generally favoring weights with large magnitudes and activations.

Based on our experiments (\cref{app:unstructured_pruning}, \cref{exp:circuit_discovery}, \cref{exp:model_correction}, and \cref{app:fig:number_of_reference_samples}), Wanda effectively identifies subgraphs relevant to model behavior with relatively few reference samples. However, as shown in \cref{app:fig:lrp_vs_wanda_sparsity}, these subgraphs are less sparse than those discovered by \gls{lrp}. In contrast, \gls{lrp} excels in discovering sparse subgraphs, as evidenced by its superior performance in the circuit discovery task (\cref{exp:circuit_discovery}) and the higher sparsity of its attribution scores (\cref{app:fig:lrp_vs_wanda_sparsity}). This method is particularly effective when a larger reference set is available, as indicated by the variability in attribution histograms (\cref{app:fig:lrp_vs_wanda_histograms}) and its gradual improvement with more reference samples (\cref{app:fig:number_of_reference_samples}). In summary, \gls{lrp} is more suitable for identifying sparse, task-relevant circuits when ample reference data is accessible. Conversely, Wanda is preferable for efficient compression when reference samples are limited.


\section{Circuit discovery}

In this section, we present the extracted \gls{ioi} circuits of the OPT model across a broad set of sparsity rates, at other granularity levels based on the description in \cref{app:prunig_approaches}. The experimental settings are similar to the configurations from \cref{exp:circuit_discovery}.


\subsection{Localization analysis for fc2}
\label{app:localzation_fc2}

\begin{figure}
  \centering
  \includegraphics[width=0.99\linewidth]{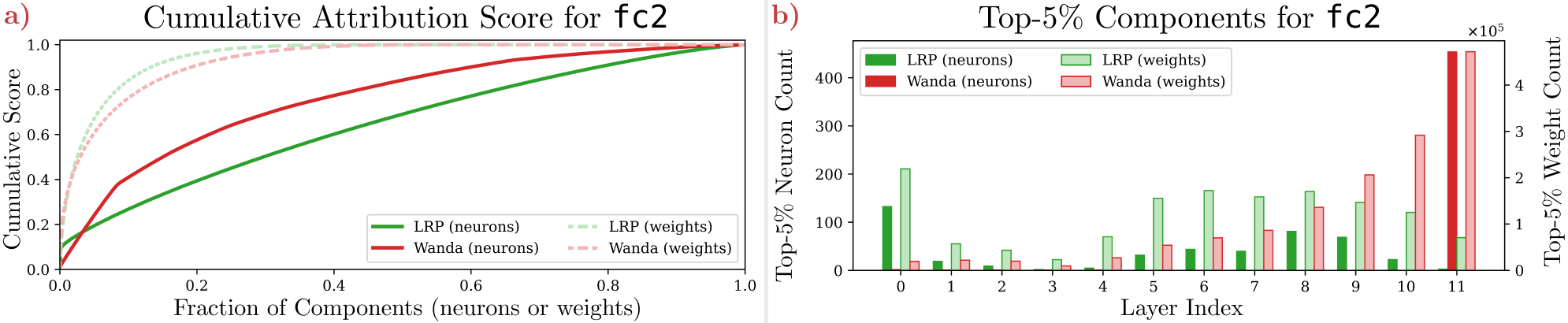}
  \caption{Localization analysis for IOI in the \texttt{fc2} layers of OPT-125M. 
  \emph{\textcolor{myred}{a)}} Cumulative relevance concentration over neuron- and weight-level components. Compared to the \texttt{fc1} analysis in the main text, \texttt{fc2} exhibits weaker and partially different localization patterns. 
  \emph{\textcolor{myred}{b)}} Layer-wise distribution of the top-5\% most relevant components. Wanda concentrates more strongly in later layers, while \gls{lrp} still identifies compact attribution structure in a subset of cases, suggesting that the clearest circuit-localization evidence arises in \texttt{fc1}..
  }
  \label{app:fig:localization_fc2}
\end{figure}

We additionally analyze attribution concentration and layer-wise localization in the \texttt{fc2} layers, shown in \cref{app:fig:localization_fc2}. Compared to \texttt{fc1}, the differences between LRP and Wanda are weaker and partially different. In particular, while LRP still yields more localized attribution at the component level in some cases, Wanda exhibits stronger concentration in the final layers. This suggests that the clearest evidence for compact multi-layer circuits arises in \texttt{fc1}, whereas \texttt{fc2} captures a complementary but less pronounced part of the behavior. We therefore focus on \texttt{fc1} in the main text and report \texttt{fc2} here for completeness.

\subsection{Circuit discovery via structured pruning}
\label{app:circuit_discovery_strcutred}

\begin{figure}
  \centering
  \includegraphics[width=0.99\linewidth]{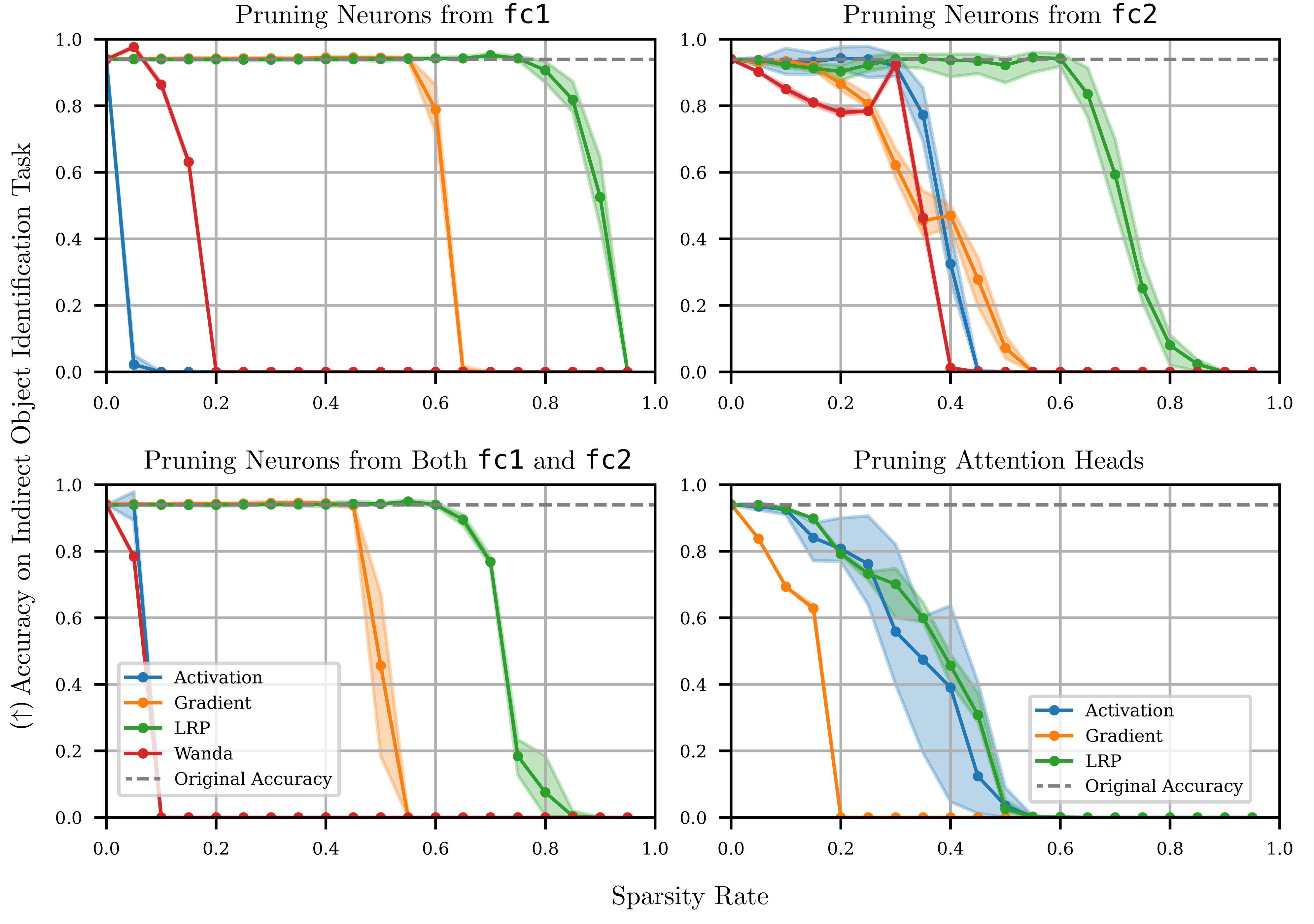}
  \caption{Based on the \gls{ioi} circuits extracted from the OPT model using structured pruning, \gls{lrp} identifies sparser and more effective circuits across neurons in \glspl{mlp} and attention heads compared Wanda and gradient. Notably, due to the absence of explicit weight parameters for individual attention heads in standard Transformer architectures, Wanda cannot be applied for circuit discovery within these heads (see \cref{app:wanda}). The shaded regions in the results represent the mean of standard deviations, reflecting the variability in circuit discovery outcomes.}
\label{fig:circuits_structured_pruning}
\end{figure}

Circuits in this scenario are derived from neurons identified via globally structured pruning (\cref{app:prunig_approaches}). As shown in \cref{fig:circuits_structured_pruning}, \gls{lrp}-extracted circuits within the \gls{mlp} layers are notably sparser and deliver superior performance compared to alternative methods. In this context, activation information was added as an additional baseline to align with \gls{acdc} \cite{conmy2023towards}. Interestingly, all methods perform poorly when targeting attention heads, underlining the crucial role of all heads working together in the \gls{ioi} task within the OPT model.

\subsection{Circuit discovery via unstructured pruning}
\label{app:circuit_discovery_unstrcutred}

\begin{figure}
  \centering
  \includegraphics[width=0.99\linewidth]{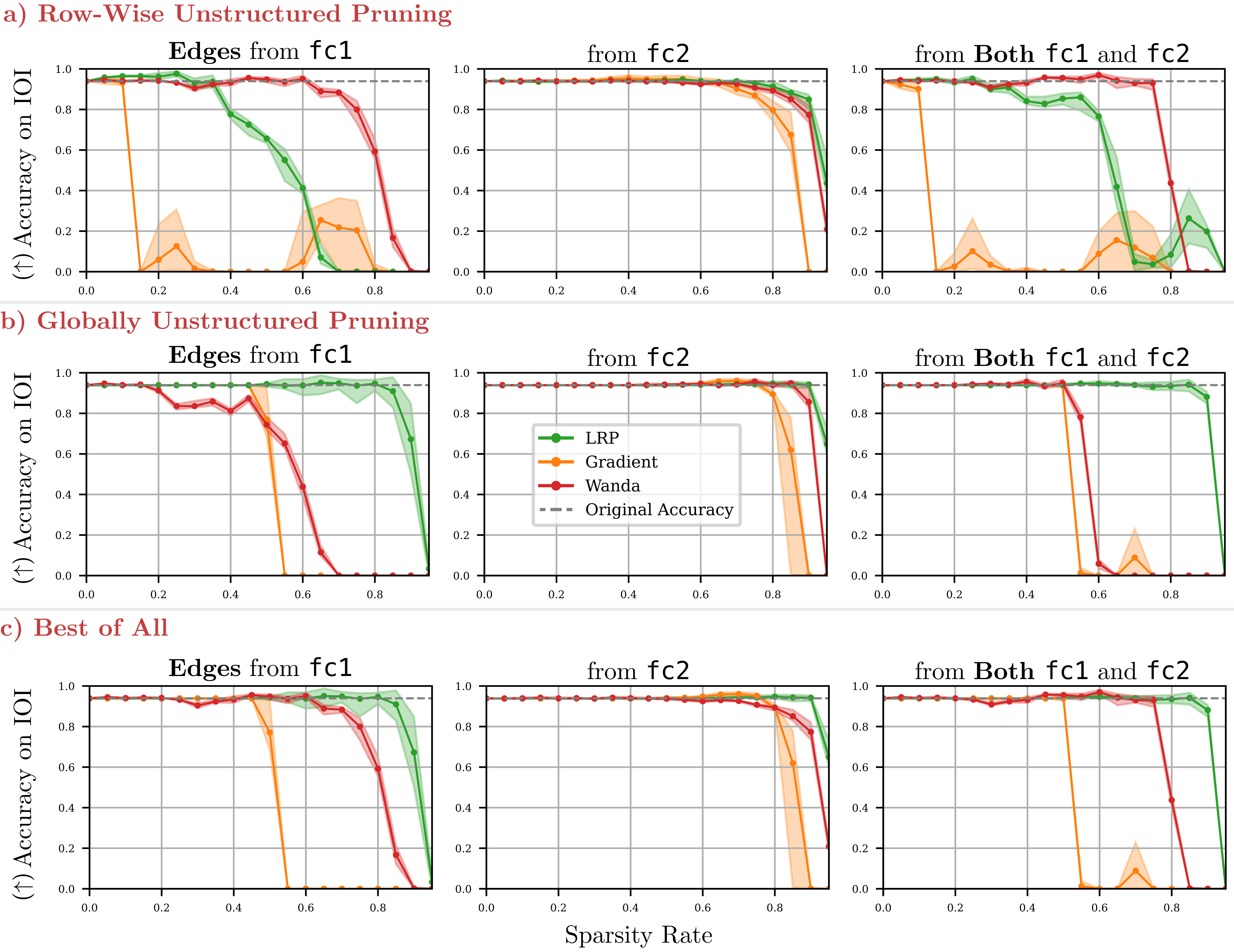}
  \caption{
  An overview of \gls{ioi} circuits discovered from the OPT model using different layer types and unstructured pruning approaches (described in \cref{app:prunig_approaches}) is shown in the figure. Panels \textcolor{myred}{a} and \textcolor{myred}{b} correspond to the row-wise and globally unstructured pruning approaches, respectively. Panel \textcolor{myred}{c} represents the best configuration for each attribution method, where the row-wise approach is optimal for Wanda and the global technique is preferred for gradient and \gls{lrp}. The shaded area in the figure indicates the mean of standard deviations.
}
  \label{fig:circuit_results_ioi_unstructured_all}
\end{figure}

In this scenario, circuits are discovered from the weight elements attributed through globally unstructured and row-wise unstructured pruning, as outlined in \cref{app:prunig_approaches}. Wanda achieves the best performance when \gls{ioi} circuits are extracted using a uniform sparsity rate applied to rows of weight elements within each linear layer, as shown in \cref{fig:circuit_results_ioi_unstructured_all}. This suggests that Wanda is particularly effective when pruning is applied uniformly at the row level. On the other hand, \gls{lrp} and gradient-based methods yield superior results when weight elements are compared across different layers, as illustrated in \cref{fig:circuit_results_ioi_unstructured_all}, indicating that these methods benefit from global pruning strategies. The optimal configurations for Wanda, \gls{lrp}, and gradient-based circuit discovery are summarized in panel \emph{a} of  \cref{fig:circuit_discovery}.

\section{Model correction}

\subsection{Toxicity improvement}
\label{app:toxicity_improvement}

\begin{figure}
  \centering
  \includegraphics[width=0.99\linewidth]{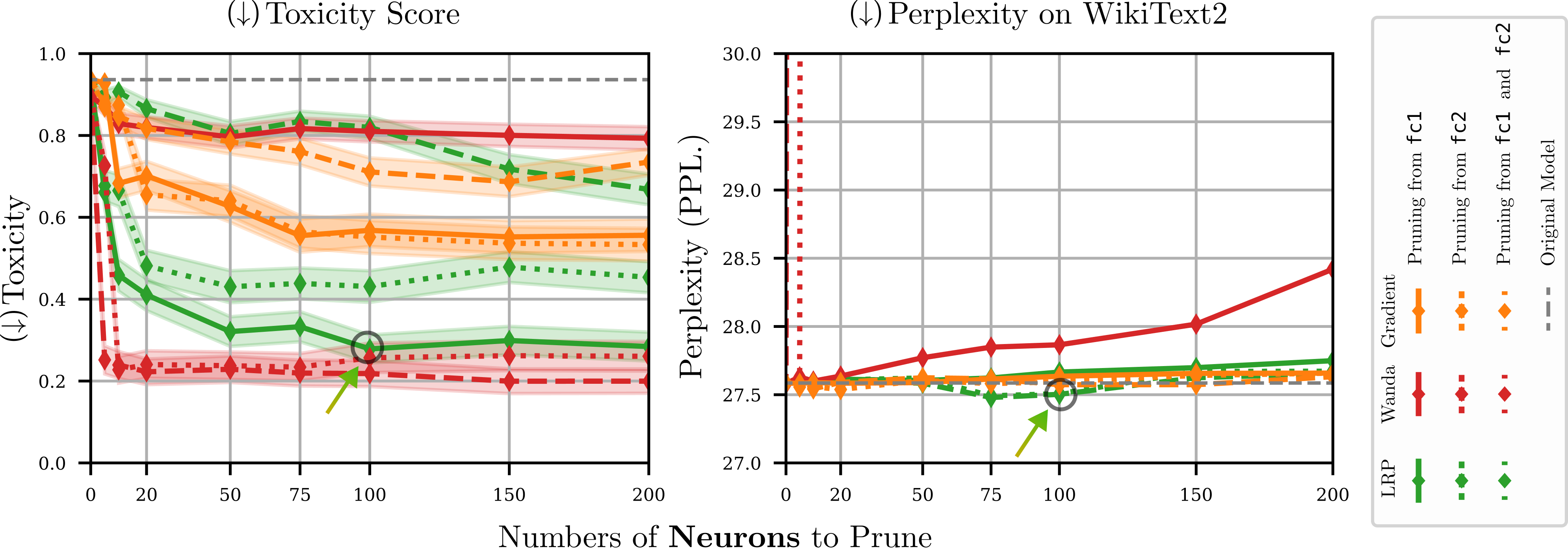}
  \caption{
  Removing neurons from different linear layers within the \glspl{mlp} blocks of the OPT model using structured pruning guided by attribution methods, effectively reduces toxicity without degrading general performance (measured by perplexity on WikiText2). Among these methods, \gls{lrp} demonstrates superior effectiveness in minimizing toxicity while preserving model accuracy. The shaded region in the figure indicates the standard error of the mean.}
  \label{fig:model_improvement_toxicity_structured_extended}
\end{figure}

\begin{figure}
  \centering
  \includegraphics[width=0.99\linewidth]{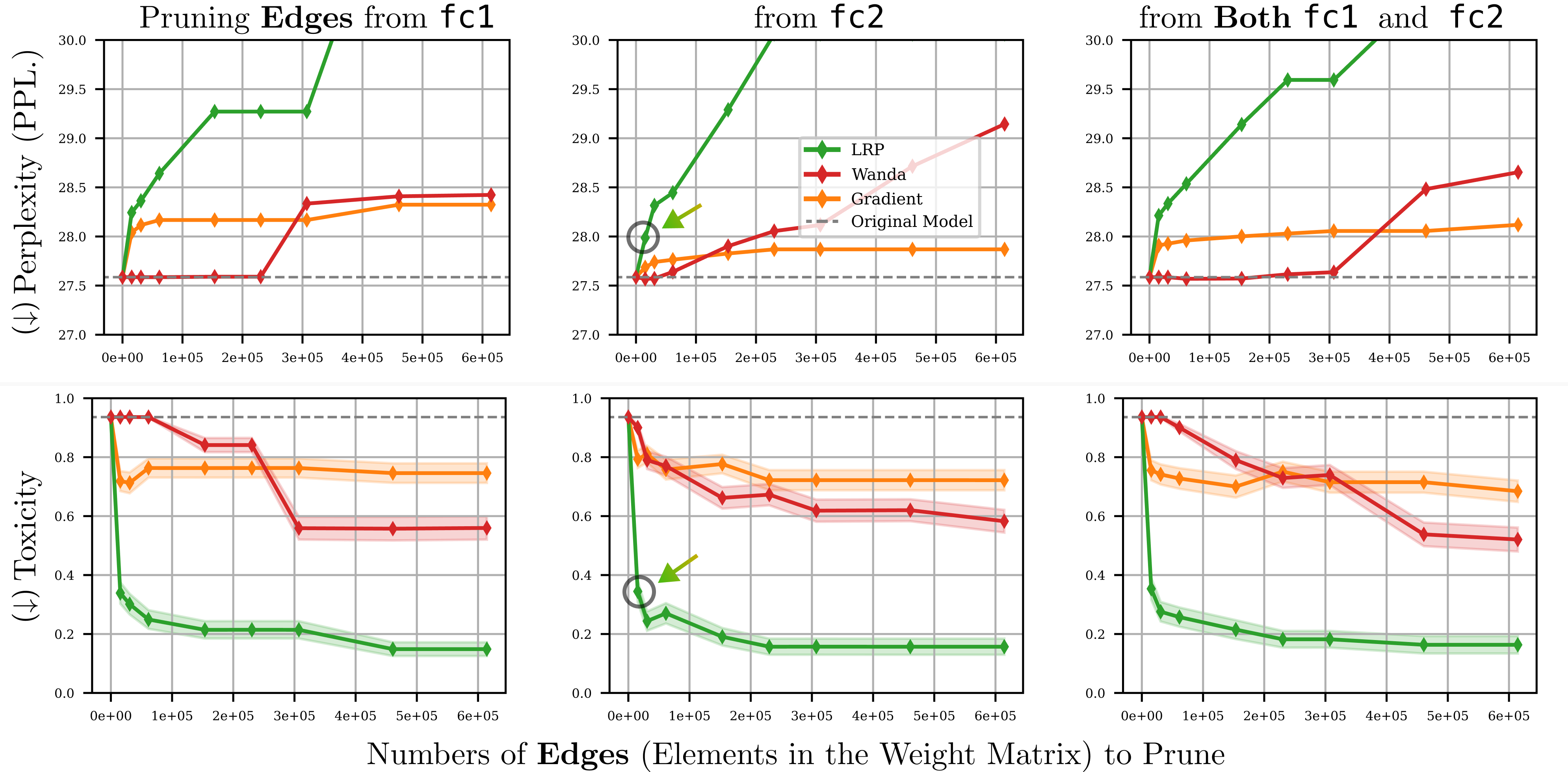}
  \caption{Pruning few weight elements across various \glspl{mlp} layers improves the toxicity score of generated responses, with methods such as \gls{lrp} showing notable effectiveness compared to Wanda and gradient. Here Wanda leverages row-wise while \gls{lrp} and gradient use global unstructured pruning.
  The shaded region in the figure indicates the standard error of the mean.}
  \label{fig:model_improvement_toxicity_unstructured_extended}
\end{figure}

As detailed in \cref{exp:model_correction}, we identified the components of the OPT model responsible for both toxic and general behaviors using \gls{lrp}, Wanda, and gradient across various granularity levels. By pruning the parameters contributing to toxic behavior, we effectively mitigated these behaviors while maintaining the model’s overall performance. The results from structured and unstructured pruning, presented in \cref{fig:model_improvement_toxicity_structured_extended} and \cref{fig:model_improvement_toxicity_unstructured_extended}, respectively, highlight the superior effectiveness of \gls{lrp} in reducing toxicity.

\subsection{Repitition improvement}
\label{app:repeatition_improvement}


\subsubsection{Repetitive response in LLMs}
As explained in \cref{exp:model_correction}, depending on the prompt and temperature setting of the \glspl{llm}, models may generate repetitive responses. These repetitions can manifest either as single tokens being repeatedly generated or as entire sequences of tokens repeating. For instance:

\begin{verbatim}
Prompt: Stop, stop, stop,
Respone:  stop, stop, stop, ...

Prompt: Love is something that
Respone:  is shared by all. Love is something that is shared by all. 

Prompt: If I repeat myself, then I repeat
Respone:  myself. I repeat myself.I repeat myself. I repeat myself.
\end{verbatim}

Such repetitive responses can be deliberately induced using specific decoding settings in the model’s generation function. Specifically, setting \texttt{temperature=0}, 
\texttt{top\texttt{\_}k=0}, and \texttt{do\texttt{\_}sample=False} ensures deterministic (greedy) decoding, which is prone to repetition:

\begin{tcolorbox}[colback=gray!5!white, colframe=gray!75!black, 
title=Generating Repetitive Responses with Hugging Face Transformers in Python, 
fonttitle=\bfseries, sharp corners, fontupper=\small]
\begin{adjustbox}{max width=\linewidth}
\begin{lstlisting}
from transformers import AutoTokenizer, AutoModelForCausalLM

tokenizer = AutoTokenizer.from_pretrained("facebook/opt-125m")
model = AutoModelForCausalLM.from_pretrained("facebook/opt-125m")

inputs = tokenizer("Love is something that", return_tensors="pt")
output = model.generate(
    **inputs,
    max_new_tokens=50,
    temperature=0.0,   # Deterministic (greedy) decoding
    top_k=0,           # No sampling, always choosing the most likely token
    do_sample=False,   # Deterministic generation
    pad_token_id=tokenizer.eos_token_id,
)
\end{lstlisting}
\end{adjustbox}
\end{tcolorbox}

These settings force the model to always select the most probable token at each step, increasing the likelihood of repetitive outputs, particularly with certain prompts.

\subsubsection{Quantifying repetitions}

Let $T = [t_1, t_2, \dots, t_n]$ represent the sequence of tokens generated by model. We define the set of unique tokens from the response as $U = \{ t_i \mid t_i \in T \}$. The \glsdesc{rur} $r$ is then calculated as:

\begin{equation}
    \centering
        r =
    \begin{cases}
    \frac{|U|}{|T|} & \text{if } |T| > 0 \\
    0 & \text{if } |T| = 0
    \end{cases}
    \label{app:eq:uniqueness}
\end{equation}


\subsubsection{Reference samples triggering repetitions}

\begin{figure}
  \centering
  \includegraphics[width=0.99\linewidth]{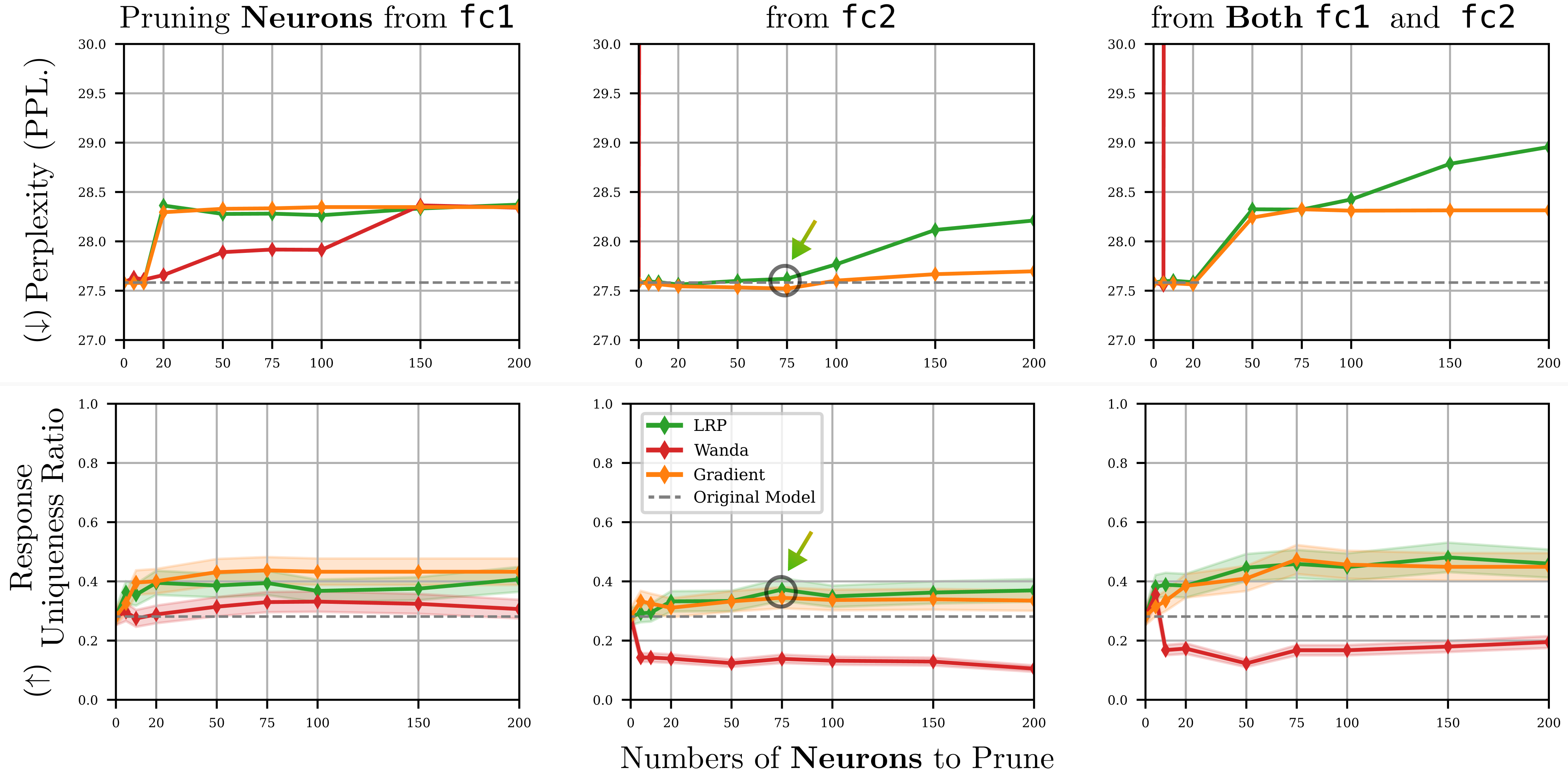}
  \caption{
  Reducing repetition in generated text can be achieved by selectively pruning neurons from linear layers, particularly \texttt{fc1} and \texttt{fc2}. Among the tested methods, removing just 20 neurons via structured pruning significantly enhances the uniqueness of responses without compromising model performance. Notably, gradient offers moderate improvements, while Wanda shows limited effectiveness in this context. Specifically, \gls{lrp} demonstrates superior performance, effectively reducing repetition with minimal pruning. The shaded area in the figure represents the standard error of the mean.}
  \label{app:fig:model_improvement_reaptition_structured}
\end{figure}

\begin{figure}
  \centering
  \includegraphics[width=0.99\linewidth]{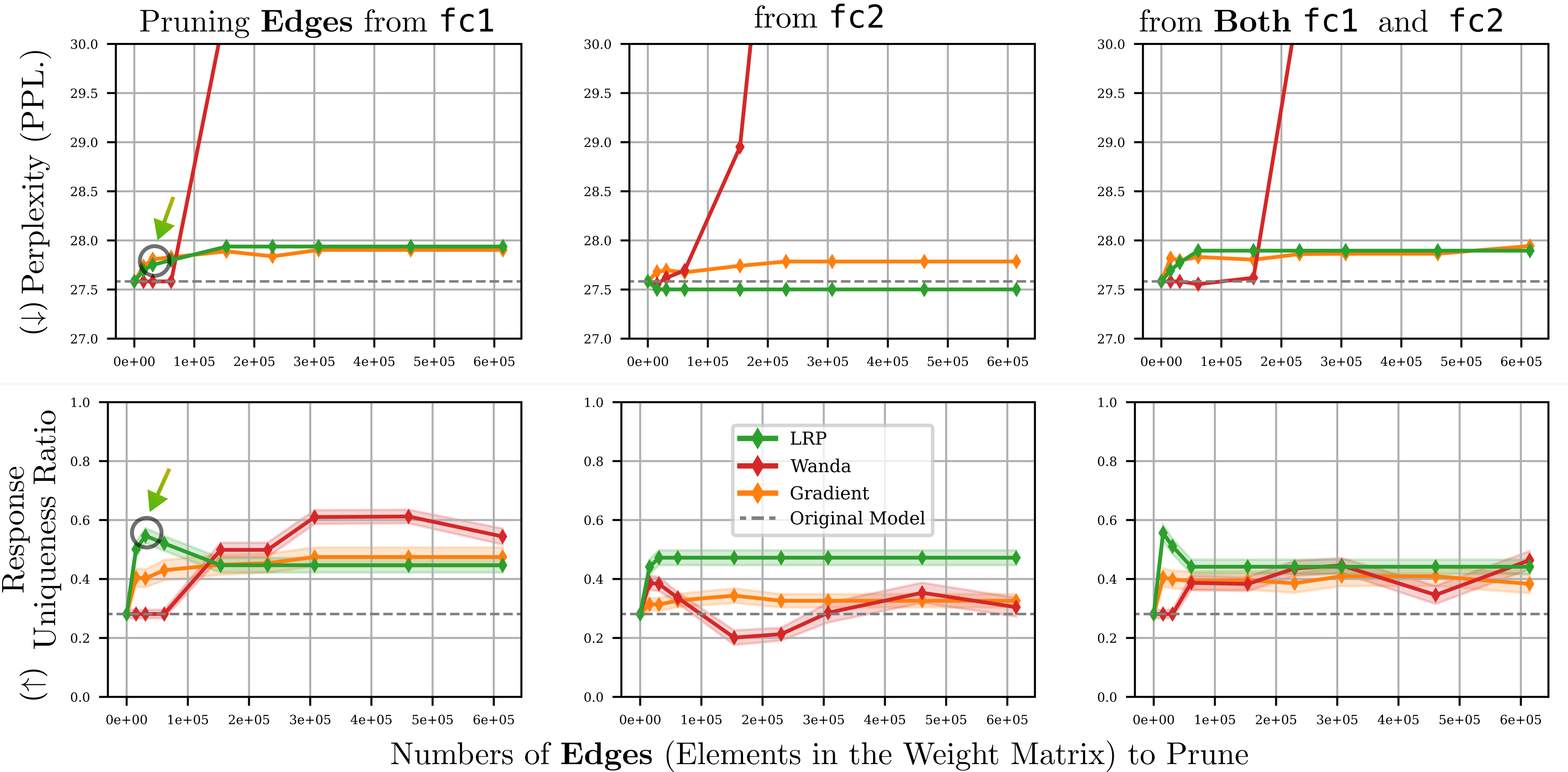}
  \caption{Following the approach in \cref{app:fig:model_improvement_reaptition_structured}, enhancing the uniqueness of generated tokens while maintaining model performance (measured by perplexity on WikiText2) can be achieved by pruning a minimal number of edges (approximately 7,000 weight elements). Among the tested methods, \gls{lrp} demonstrates notable effectiveness, achieving these improvements with minimal sparsity rates. 
  Here Wanda leverages row-wise while \gls{lrp} and gradient use global unstructured pruning.
  The shaded area in the figure represents the standard error of the mean.}
  \label{app:fig:model_improvement_reaptition_unstructured}
\end{figure}

We asked ChatGPT to generate a set of prompts that lead to repetitive responses, which we then selected 53 samples from, each characterized by a low \glsdesc{rur} ($r < 0.5$). These prompts, detailed in \cref{app:table:repeatitive_prompts}, constitute the set $\mathcal{X}_{\text{ref}}^{\text{Undesired}}$, used for attribution. For comparison, a general set $\mathcal{X}_{\text{ref}}^{\text{General}}$ was created following \cref{exp:model_correction}. Our correction method (\cref{sec:methods:model_correction}) effectively mitigates repetition, as shown in \cref{app:fig:model_improvement_reaptition_structured} and \cref{app:fig:model_improvement_reaptition_unstructured}, using structured and unstructured pruning, respectively. Notably, \gls{lrp} achieves superior performance, reducing repetition while preserving model perplexity on WikiText2. This improvement is attained with fewer pruned parameters than other methods. Given the substantial gains, we recommend unstructured pruning for optimal repetition reduction.

\begin{longtable}{>{\ttfamily\raggedright\arraybackslash}p{0.3\linewidth} >{\ttfamily\raggedright\arraybackslash}p{0.3\linewidth} >{\ttfamily\raggedright\arraybackslash}p{0.3\linewidth}}
\caption{List of 53 prompts that trigger highly repetitive responses for the 125M-parameter of the OPT model, each resulting in a \glsdesc{rur} of $r < 0.5$.} \label{app:table:repeatitive_prompts} \\
\toprule
``Happiness can be found in'' & ``If I had one wish, it would be'' & ``Love is something that'' \\
``There is no doubt that'' & ``I wake up every day because'' & ``Sometimes, I wonder if'' \\
``The secret to success is'' & ``Fate is what we make of'' & ``Once upon a time'' \\
``There was a boy who'' & ``In a magical forest, there lived'' & ``Long ago in a distant land'' \\
``In the middle of nowhere'' & ``The princess whispered to'' & ``It was the beginning of the end when'' \\
``The wizard cast a spell and'' & ``Suddenly, the ground shook and'' & ``Keep saying it:'' \\
``I told you to repeat:'' & ``Repeat this:'' & ``Again and again I say'' \\
``Copy this:'' & ``Echo these words:'' & ``Repeat. Repeat. Repeat.'' \\
``I’ll say it again:'' & ``Say it one more time:'' & ``Can you say this again:'' \\
``Repeat this forever:'' & ``Things I like: apples, bananas,'' & ``My top five choices are'' \\
``These are my favorites:'' & ``Consider the sequence:'' & ``The next on the list is:'' \\
``Well, well, well'' & ``So, so, so'' & ``Like, like, like'' \\
``Okay, okay, okay'' & ``Um, um, um'' & ``Hmm, hmm, hmm'' \\
``Ah, ah, ah'' & ``Alright, alright, alright'' & ``Really, really, really'' \\
``Fine, fine, fine'' & ``Maybe, maybe, maybe'' & ``Hey, hey, hey'' \\
``No, no, no'' & ``Stop, stop, stop'' & ``Listen, listen, listen'' \\
``Now, now, now'' & ``I am what I am'' & ``You are who you are'' \\
``If I repeat myself, then I repeat'' & ``There is no end to this'' & ``And then, and then, and then'' \\
``Talking about talking'' & ``Explaining an explanation'' & \\
\bottomrule
\end{longtable}

\section{Generalization on Additional Models (Qwen2-0.5B and TinyLlama)}
\label{app:sec:other_models}

\subsubsection{Experimental Setup}
We extend our evaluation to additional architectures, namely Qwen2-0.5B and TinyLlama-1B, to assess the generalizability of attribution-guided pruning beyond OPT-125M considered in the main text. We follow the same experimental protocol, including identical pruning procedures, evaluation metrics, and datasets for IOI, repetition, toxicity, and perplexity. Unless otherwise specified, all hyperparameters and sparsity schedules are kept unchanged. For these models, we focus on linear layers within the \gls{mlp} blocks. In contrast to OPT-125M, where MLPs consist of \texttt{fc1} and \texttt{fc2}, Qwen2 and TinyLlama employ a gated MLP architecture with three projections: \texttt{down\_proj}, \texttt{up\_proj}, and \texttt{gate\_proj}. We evaluate all three components to assess whether attribution-guided pruning generalizes across architectural variants. We report results for \gls{lrp}, Wanda, and gradient-based attribution in this section to study the effect of attribution signals across architectures.


\subsection{Circuit Discovery}
\label{app:sec:other_models:circuit_discovery}

We evaluate circuit discovery on the IOI task under both structured (neuron-level) and unstructured (weight-level) pruning. Attribution scores are computed on 128 samples, and evaluation is performed on 512 samples, consistent with the main experiments.

\begin{figure}
  \centering
  \includegraphics[width=0.99\linewidth]{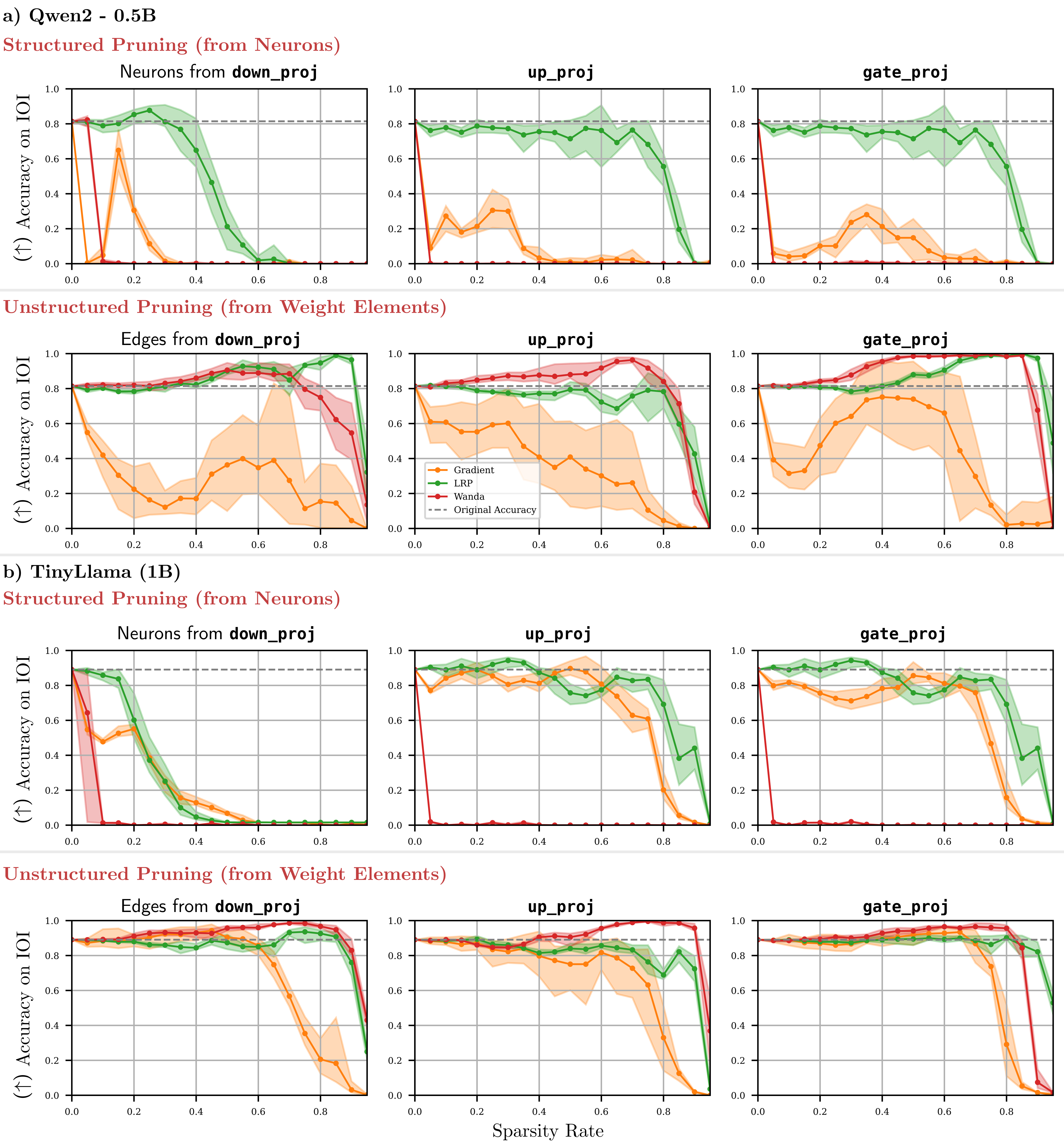}
  \caption{Circuit discovery performance on the IOI task for Qwen2-0.5B and TinyLlama-1B under structured (\emph{top}) and unstructured (\emph{bottom}) pruning. \gls{lrp} consistently preserves task accuracy at higher sparsity levels compared to Wanda and gradient-based pruning, particularly in structured settings. In unstructured pruning, \gls{lrp} alongside Wanda (via row-wise usntrcutred pruning) maintain stable performance across sparsity levels, while gradient exhibits higher variance and earlier degradation.
Shaded regions indicate variability across seeds.
}
    \label{app:figure:circuit_discovery_qwen_tinyllama}
\end{figure}

As shown in \cref{app:figure:circuit_discovery_qwen_tinyllama}, \gls{lrp} consistently preserves IOI accuracy at substantially higher sparsity levels compared to gradient-based pruning across both Qwen2 and TinyLlama. In structured pruning, \gls{lrp} maintains near-original performance over a wide sparsity range, indicating effective identification of neuron-level circuits, while gradient-based pruning rapidly degrades. In unstructured pruning, \gls{lrp} continues to enable highly sparse subnetworks with stable performance, whereas gradient-based pruning remains both less stable and more sensitive to sparsity, as reflected by wider confidence intervals. While forward-pass-based methods such as Wanda can perform competitively under specific pruning configurations (\eg row-wise unstructured pruning), our results focus on attribution-based comparisons to isolate the effect of attribution quality across architectures.


\paragraph{Conclusion.}

These results demonstrate that attribution-guided circuit discovery via \gls{lrp} generalizes across architectures and remains effective under both structured and unstructured pruning. In particular, \gls{lrp} consistently identifies compact and behaviorally relevant components, while providing more stable and controllable sparsity-performance trade-offs than gradient-based attribution.

\subsection{Model Correction}
\label{app:sec:other_models:model_correction}

We evaluate attribution-guided pruning for targeted model correction on repetition and toxicity. 
For toxicity, we select prompts from the RealToxicityPrompts dataset with scores above $0.9$, resulting in 218 samples for Qwen2-0.5B and 95 samples for TinyLlama. For repetition, we select prompts with \gls{rur} scores below $0.5$, yielding 159 and 120 samples, respectively. For structured pruning, we sweep sparsity levels corresponding to approximately $5$ to $5000$ neurons removed, covering a broad range of pruning regimes. Unstructured pruning follows an analogous dense-to-sparse schedule.

\begin{figure}
  \centering
  \includegraphics[width=0.99\linewidth]{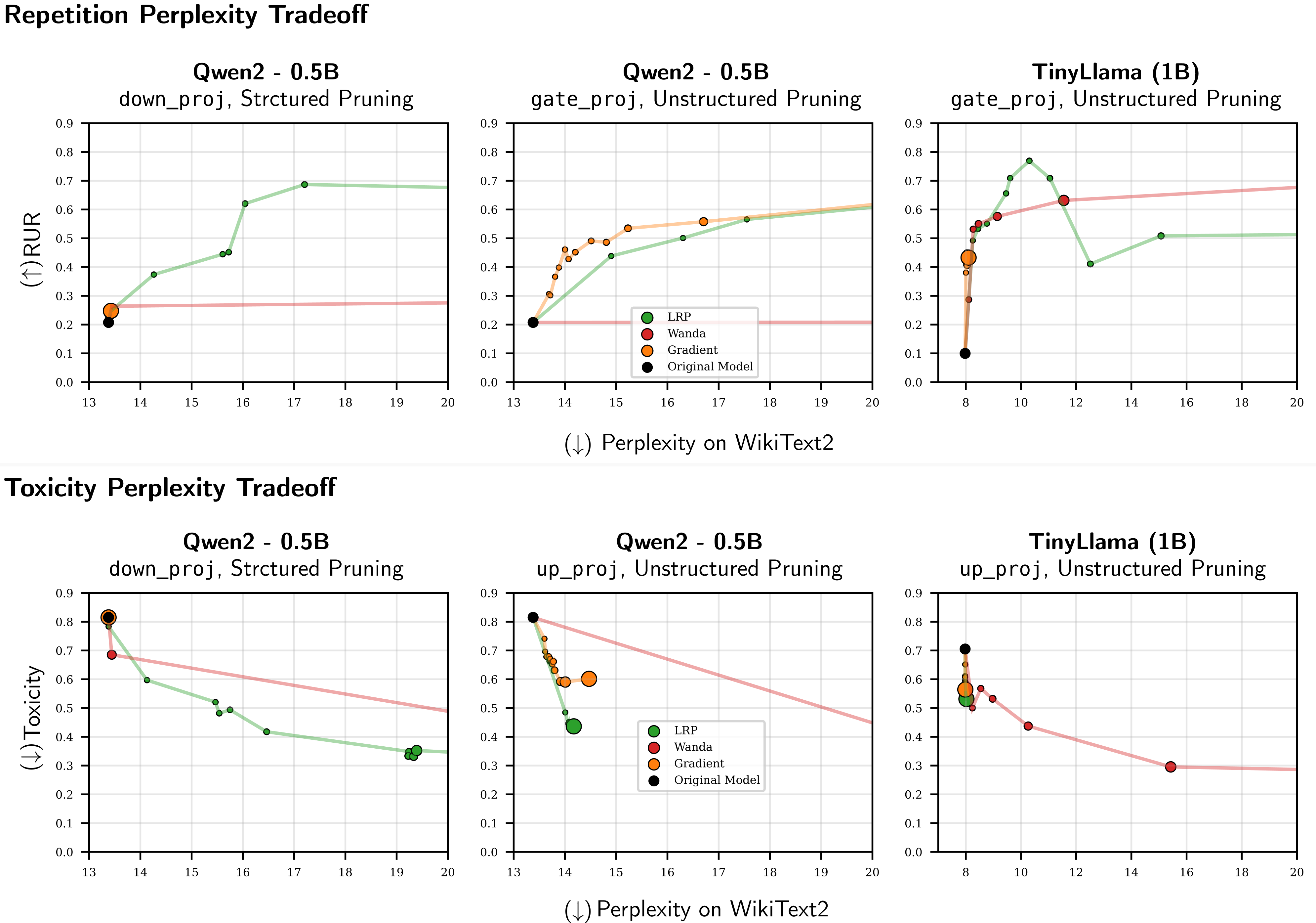}
  \caption{Trade-offs between perplexity and behavioral metrics (repetition and toxicity) for Qwen2-0.5B and TinyLlama-1B. \gls{lrp} achieves improved trade-offs, reducing undesirable behaviors at comparable or lower perplexity compared to gradient-based pruning and Wanda. Marker size indicates sparsity level. In several cases (\eg Qwen2 structured pruning), gradient results collapse to a narrow range of operating points, while Wanda exhibits less consistent trade-offs, highlighting the improved controllability of \gls{lrp}.}
    \label{app:figure:model_correction}
\end{figure}

\cref{app:figure:model_correction} shows the trade-off between perplexity and behavioral metrics across pruning configurations. Across both models and tasks, \gls{lrp} consistently achieves more favorable trade-offs, reducing repetition and toxicity at comparable or lower perplexity than gradient and Wanda. In several settings (\eg Qwen2 structured pruning), gradient collapses to a narrow set of operating points, where increasing sparsity produces little to no change in model behavior, indicating limited ability to rank components by behavioral relevance. Wanda, while effective for general sparsification, exhibits less consistent behavior for targeted correction (similarly shown in \cref{app:figure:model_correction}), often requiring larger perplexity degradation to achieve comparable reductions in undesired behavior. In contrast, \gls{lrp} produces well-structured trade-off curves, enabling smooth and controllable transitions between model quality and behavioral correction.


\paragraph{Conclusion.}
These results show that \gls{lrp}-based attribution enables more effective and controllable model correction across architectures, consistently outperforming gradient-based pruning and providing more reliable trade-offs than Wanda for behavior-specific interventions.

\section{Used resources}
\label{app:used_resources}

All experiments were conducted on NVIDIA A100 GPUs (40GB). Both \gls{lrp} and Wanda are efficient, with \gls{lrp} requiring a forward and backward pass for attribution ($\approx$5 seconds per reference sample with 2048 tokens on TinyLlama), while Wanda only uses forward passes ($\approx$1 second per sample). As model size or reference set grows, their computation times increase.

Our implementation of \gls{lrp} follows an explicit formulation, similar to prior implementations such as Zennit~\cite{anders2021software}, where activations and relevance scores are stored during forward and backward passes. This results in higher memory consumption: approximately 60GB of VRAM is required for attributing a single 2048-token sequence on TinyLlama. In contrast, Wanda requires approximately 5GB of VRAM, although its memory usage increases with large reference sets, as noted in its official implementation. We note that this memory overhead is not inherent to \gls{lrp}. More memory-efficient implementations based on implicit formulations (\eg \cite{rezaei2024mambalrp, arras2025close}) avoid storing intermediate relevance tensors, significantly reducing GPU memory requirements. Therefore, the reported memory usage reflects our explicit implementation choice rather than a fundamental limitation of \gls{lrp}. In both methods, memory usage scales with sequence length and the size of the reference set.


\clearpage

\end{document}